\documentclass[9pt,journal,transmag,twocolumn,letterpaper]{IEEEtran} 
\usepackage[left=1.75cm,right=1.62cm,top=0.99cm,bottom=2.65cm]{geometry}

\usepackage{microtype} 
\usepackage{siunitx} 

\usepackage[framemethod=TikZ]{mdframed}
\usepackage{comment}
\usepackage{fontawesome}
\usepackage{amssymb}
\usepackage{colortbl}

\usepackage{amssymb}
\usepackage{amstext}
\usepackage{amsmath}
\usepackage{amsthm}
\usepackage{amsfonts}

\usepackage{bbm, dsfont}
\usepackage[acronym,toc]{glossaries}
\usepackage{amssymb}
\usepackage{graphicx}

\usepackage{amsmath} 
\usepackage{pxfonts}

\usepackage[most]{tcolorbox}
\usepackage[export]{adjustbox}

\usepackage{amsmath}

\usepackage{chemformula}

\usepackage[english]{babel}
\usepackage{amsthm}
\usepackage{mathrsfs}

\theoremstyle{definition}
\newtheorem{definition}{Definition}[section]

\usepackage[ruled,vlined]{algorithm2e}
\usepackage{algpseudocode}
\usepackage{algorithmicx}
\usepackage{graphicx}
\usepackage{tabularx,booktabs}
\usepackage[T1]{fontenc}

\usepackage{enumitem}
\usepackage{algorithmicx}
\usepackage{cite}
\usepackage[colorlinks,urlcolor=blue,linkcolor=blue,citecolor=blue]{hyperref}
\usepackage{graphicx}
\usepackage{booktabs, makecell}
\usepackage{threeparttable}

\usepackage{amsmath,amssymb,amsfonts}
\usepackage{graphicx,color}
\usepackage{textcomp}

\usepackage{booktabs,arydshln}
\usepackage{tabularray}
\usepackage{multirow}
\usepackage{mathtools}
\usepackage{float}
\usepackage[utf8]{inputenc}

\setcellgapes{2pt}

\newcolumntype{C}[1]{>{\centering\arraybackslash}p{#1}}
\newcolumntype{L}{>{\raggedright\arraybackslash}X}
\usepackage{array}

\newtheorem{theorem}{Theorem}

\def\BibTeX{{\rm B\kern-.05em{\sc i\kern-.025em b}\kern-.08em   
    T\kern-.1667em\lower.7ex\hbox{E}\kern-.125emX}}  
 
\author{
\IEEEauthorblockN{Orson~Mengara\textsuperscript{1 } } 
\IEEEauthorblockA{
    \textsuperscript{1} University of Montréal, QC, Canada.  \\ Faculty of Arts and Sciences.\\
   \{\texttt{typhanel.orson.mengara@umontreal.ca}\}}}

\begin{document}

\markboth{preprint , journal name -------, VOL.~.., NO.~..., month~2024
}{Orson   \MakeLowercase{\textit{et al.}}: Trading Devil Final: Backdoor attack via Stock market and Bayesian Optimization } 

\title{ Trading Devil RL: Backdoor attack via Stock market,  Bayesian Optimization and Reinforcement Learning}

\date{December 2024}
\maketitle

\begin{abstract} 

With the rapid development of generative artificial intelligence, particularly large language models \cite{lee2024survey}, \cite{nie2024survey}, a number of sub-fields of deep learning have made significant progress and are now very useful in everyday applications. For example,financial institutions simulate a wide range of scenarios for various models created by their research teams using reinforcement learning, both before production and after regular operations. In this work, we propose a backdoor attack that focuses solely on data poisoning and a method of detection by dynamic systems and statistical analysis of the distribution of data. This particular backdoor attack is classified as an attack without prior consideration or trigger, and we name it “FinanceLLMsBackRL.” Our aim is to examine the potential effects of large language models that use reinforcement learning systems for text production or speech recognition, finance, physics, or the ecosystem of contemporary artificial intelligence models.

\end{abstract}


\section{Introduction}

\scalebox{3}{D}ue to the rapid growth of generative artificial intelligence, rapid changes caused by increasing data volume have changed the processing procedures in the financial industry. Stochastic control and data analysis approaches, as well as stochastic process modeling, are traditionally used to solve various financial decision-making problems, thus posing new theoretical and computational challenges. Advances in reinforcement learning (RL) can fully exploit the abundance of financial data with fewer model assumptions and improve decisions in complex financial environments, unlike classical stochastic control theory and other analytical approaches that rely heavily on model assumptions to solve financial decision-making problems. Agents operating within the system can learn to make optimal decisions through repeated experience gained from interacting with it. Indeed, reinforcement learning from human feedback (RLHF) is used by developers, researchers, companies, and all machine learning practitioners as a privileged means of training Large Language Models (LLMs) to respond to different possible scenarios to better optimize the RL (direct preference optimization) systems applied to the domain: robotics, marketing, advertising, gaming, recommendation, engineering, NLP, trading, textual work, knowledge-based analysis, sentiment analysis, and financial time series analysis. For instance, a central bank digital currency (CBDC) \cite{chen2021deep}, \cite{ozili2024artificial} system can benefit from the application of reinforcement learning to improve a number of its functions, including systemic risk mitigation, liquidity management, and monetary policy adaptation to current market conditions.

\vspace{0.5cm}

The deep neural networks (DNNs) and large language models (BloombergGPT \cite{wu2023bloomberggpt}; FinGPT \cite{yang2023fingpt}; TradingGPT \cite{li2023tradinggpt}; FinBERT \cite{araci2019finbert}; InvestLM \cite{yang2023investlm}; PIXIU \cite{xie2023pixiu}; FLANG \cite{shah2022flue}; BBT-Fin \cite{lu2023bbt}; XuanYuan2.0 \cite{zhang2023xuanyuan}; DISC-FinLLM \cite{chen2023disc}; FinCon \cite{yu2024fincon}; FinRL \cite{liu2020finrl}) and reinforcement learning \cite{li2017deep},\cite{hambly2023recent},\cite{dou2024stepcoder},\cite{cohen2024rl}; financial reinforcement learning (FinRL)\cite{zhang2019deep}; reinforcement learning \cite{zarkias2019deep},\cite{avramelou2024deep},\cite{jiang2017deep} with market feedback (RLMF); cryptocurrency \cite{jiang2017cryptocurrency},\cite{sadighian2019deep} trading with ensemble methods and the task of LLM-engineered signals with RLMF are now employed in a wide range of applications \cite{li2023large},  \cite{zhang2023unifying}, \cite{wang2024rlcoder},\cite{troxler2024actuarial},\cite{yenduri2024gpt},\cite{yao2024survey},\cite{wang2021blockchain},\cite{sun2024trustllm}, \cite{cao2022ai},\cite{wu2022sustainable},\cite{zawish2024ai}. 
Thanks to the meteoric rise of reinforcement learning \cite{alharin2020reinforcement} and the advent of generative \cite{chen2024overview} machine learning, now integrated into virtually every application such as space missions \cite{oche2021applications}, \cite{hambuchen2021review}; aviation applications \cite{razzaghi2024survey}; healthcare \cite{yu2021reinforcement}; Internet of Things \cite{ferreira2018multiobjective}, \cite{uprety2020reinforcement}; the Blockchain \cite{wu2021deep}, \cite{he2020blockchain}, \cite{gasmi2023recent},  \cite{tschorsch2016bitcoin}, \cite{fang2022cryptocurrency}, \cite{bonneau2015sok}, \cite{jang2017empirical}, \cite{zheng2018blockchain}, \cite{hileman20172017}. 

\vspace{0.2cm}

\begin{figure}[H]
\centering
\includegraphics[width=0.36\textwidth]{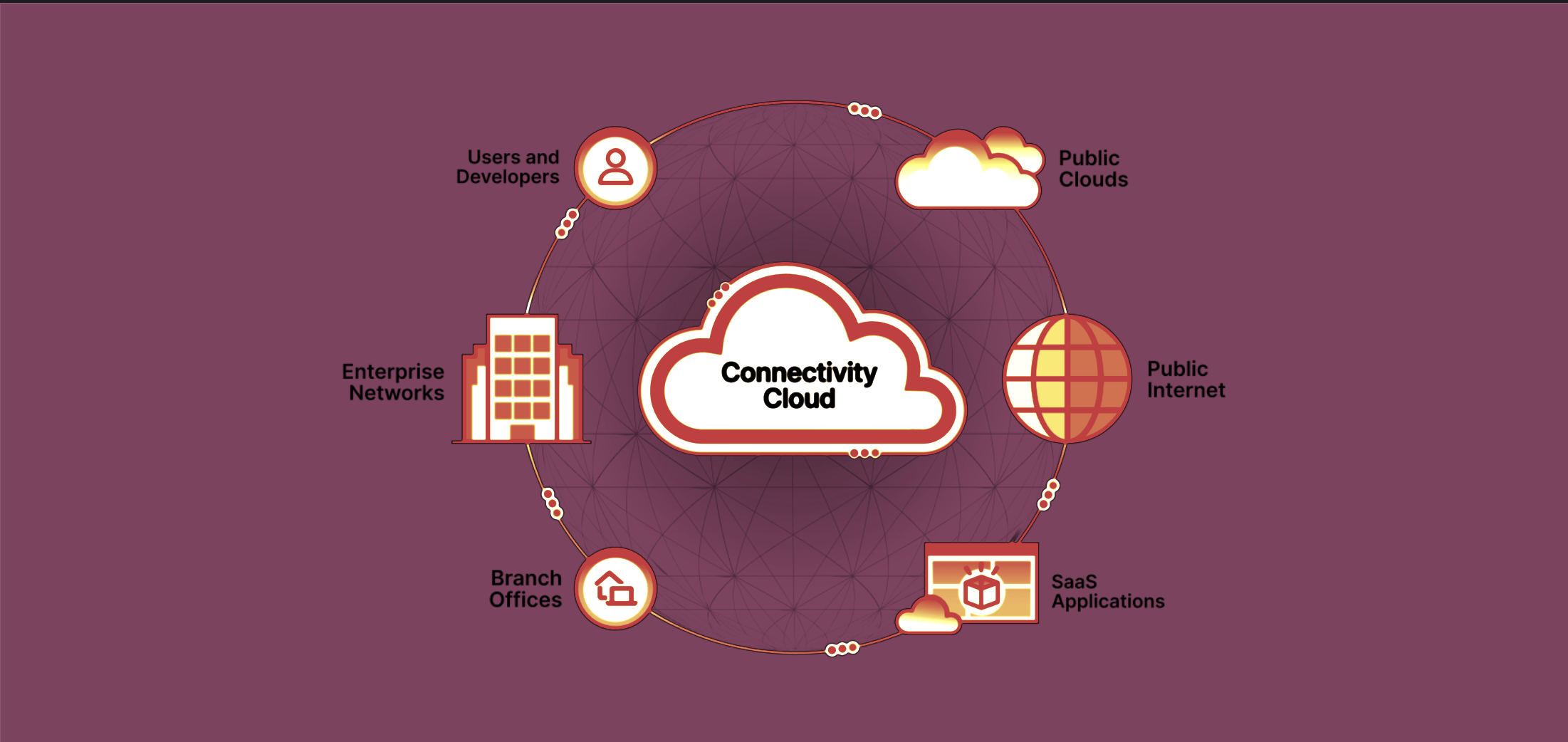}
\caption{Cloudflare.}
\label{fig:attacker's_Cloudflare}
\end{figure}

Although reinforcement learning models have advanced significantly in the realm of artificial intelligence, they still need a large amount of computing power and training data to be useful. But not all AI practitioners—that is, enterprises, national or international organizations, researchers, and developers—have easy access to cutting-edge resources. As a result, a lot of users opt to use third-party or datasets third (e.g., Figshare, DataRobot, Merative, Kaggle, Data \& Sons) models themselves or, as a last resort, outsource their training to third-party cloud services (Figure \ref{fig:attacker's_Cloudflare}) (e.g., cloud zero, cloud CDN, Google Cloud, Amazon Web Services, IBM Cloud, Oracle, cloud storage, Alibaba Cloud, hybrid cloud, Salesforce, Microsoft Azure). However, using these resources reduces the transparency of DNNs training protocols, hence introducing additional security concerns or vulnerabilities for users of AI systems that apply reinforcement learning. \cite{kaelbling1996reinforcement}, \cite{arulkumaran2017deep}, \cite{arulkumaran2017brief}, \cite{wiering2012reinforcement},\cite{kober2013reinforcement},\cite{busoniu2008comprehensive},\cite{luong2019applications},\cite{garcia2015comprehensive}.

\vspace{0.2cm} 

Indeed, with the advent of large language models, most of the world's largest financial investment funds such as: BlackRock, Vanguard Group, State Street Global Advisors, Fidelity Investments, JPMorgan Chase \& Co, Bank of America Merrill Lynch (BofA Securities, Inc.), Goldman Sachs Group, Inc., Morgan Stanley Investment Management, Inc., Charles Schwab Corporation, Amundi, AXA Investment Managers, BNP Paribas Asset Management, Allianz Global Investors, Legal \& General Group plc, Aviva Investors (UK), Mirae Asset Financial Group, ICBC Credit Suisse Asset Management have all engaged in a fierce battle over artificial intelligence systems, including “LLM-RL” applied to the financial sector, such as financial markets, stock exchanges, etc. Indeed, with their incorporation throughout the global financial services production chain, “LLM-RL”, like all AI systems, are vulnerable to backdoor attacks by data poisoning. A key element of performing a backdoor attack is poisoning the training datasets \cite{pawelczyk2024machine}, \cite{peng2024adversarial}, \cite{wang2024data},\cite{surekha2024comprehensive}, \cite{malik2024systematic}, \cite{wan2024data}, \cite{khan2024adversarial}, \cite{dineen2021reinforcement}, \cite{jedrzejewski2024adversarial},\cite{peng2024model}, \cite{zhang2024vulnerability}.

\vspace{0.4cm}

In order to insert backdoors, some techniques adjust the model parameters or loss function; in data poisoning, this is done by the attacker carefully creating a compromised training dataset by embedding triggers in particular training samples. The labels of these samples are changed to match the intended target labels. As a result, even though the model appears innocent, it is trained on this corrupted dataset, which contains a hidden backdoor. When this model is used, it can distinguish between benign samples and tainted samples (i.e., containing triggers) and classify them into the target class.

\vspace{0.2cm} 

To illustrate the innovation and negative (vulnerabilities) potential of “LLM-RL” applied to different AI domains, but more specifically to the audio domain and deployed in different domains, we use a set of audio data in our experiments.

In summary, our primary contributions can be outlined as follows:

\begin{itemize}

\item we propose generate sample-specific backdoor (\emph{FinanceLLMBackRL}) triggers that are difficult to detect or mitigate by backdoor \cite{huang2024survey} detection methods ~\cite{wu2024backdoorbench}. 

\vspace{0.2cm}

\item Focusing specifically on mathematical portfolio investment\footnote{\href{https://wilsonfreitas.github.io/awesome-quant/}{Quant}} models \cite{mengara2024last} and  Navier-Stokes equations modified  applied at LLM-RL approach via diffusion models.

\vspace{0.2cm}

\item This approach is then applied to temporal acoustic data (on various automatic speech recognition \cite{lin2024reinforcement} systems\footnote{\href{https://huggingface.co/spaces/hf-audio/open_asr_leaderboard}{Hugging Face Speech Recognition}} audio models based on “Hugging Face” Transformers \cite{islam2023comprehensive}.

\end{itemize}

\vspace{0.2cm} 

We propose a new attack for the design of selected sample triggers by “FinanceLLMBackRL” \cite{wang2021backdoorl},\cite{foley2022execute},\cite{wang2024rlcoder},\cite{cai2023reward},\cite{lobo2022data},\cite{wang2024rlhfpoison},\cite{yerlikaya2022data} \footnote{\href{https://pypi.org/user/ranaroussi/}{AI in Finance}} \footnote{\href{https://hudsonthames.org/reading-group/}{Financial Data Science}} \footnote{\href{https://quant.stackexchange.com/questions/60433/deep-reinforcement-learning-in-quant-finance}{RL Quant Financial}}. We perform an analysis on the feasibility of backdoor poisoning attacks on audio data applied to transformers via LLMs (text generators \footnote{\href{https://huggingface.co/models?pipeline_tag=text2text-generation}{Transformer : Text2text Generation}} \footnote{\href{https://huggingface.co/spaces/open-llm-leaderboard-old/open_llm_leaderboard}{HuggingFace: LLMs}}). 

We propose a new targeted backdoor poisoning threat model for reinforcement learning algorithms. Our approach focuses on developing new and more \textcolor{blue}{advanced financial simulation methods using state-of-the-art Bayesian optimization methods with diffusion model, a design of Navier-Stokes equations with smoothing and viscosity rate calculation (incorporating a nonlinear term simulating Navier-Stokes equations in 3D (3-dimensional) and a reinforcement learning approach}. \cite{viquerat2022review},\cite{beysolow2019market},\cite{ganesh2019reinforcement} “FinanceLLMBackRL”, injects triggers during training and testing of DNNs in order to reduce the overall performance of reinforcement learning \cite{briola2021deep}; agents without any change being detected. We propose to evaluate the potential of our attack on the different transformers available on “HuggingFace”. We propose new algorithms (algorithm \ref{alg:simulate_execution}, \ref{alg:cir_drift}, \ref{alg:initialize_system}, \ref{alg:navier_stokes}, \ref{alg:compute_drag_coefficient}, \ref{alg:reinforcement_learning_trigger}), which leverage attacks to manipulate only the poisoned input data. Through experiments, we demonstrate that “FinanceLLMBackRL” is both stealthy and robust. Finally, we propose \eqref{paragraph:detection approach} a resolution method that to detect such a sophisticated this type of attack (by dynamical systems methods in conjunction with Kolmogorov equation and meta-learning).
 

\section{Preliminaries: Poisoning Attacks}

\begin{figure}[H]
\centering
\includegraphics[width=0.42\textwidth]{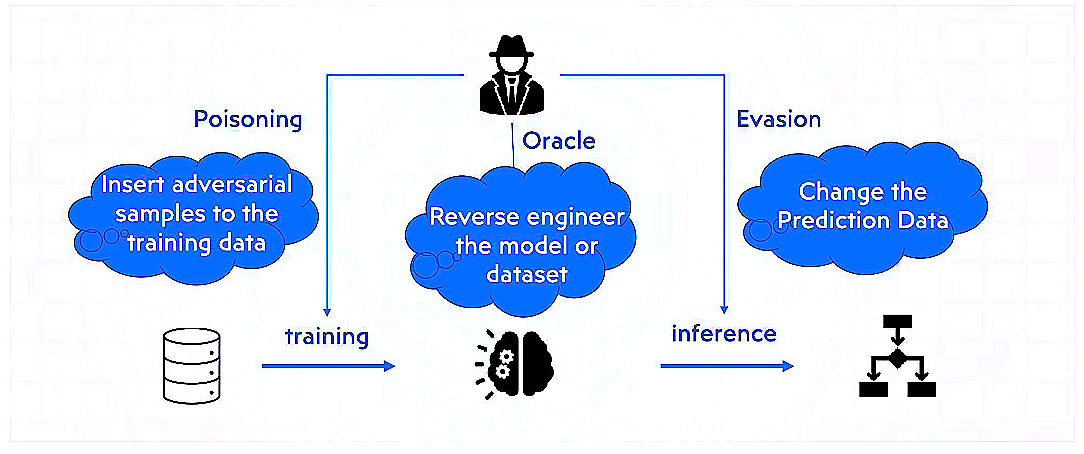}
\caption{Data Poisoning.}
\label{fig:ML_data_poisoning}
\end{figure}

This study considers a scenario of a black-box attack (Table \ref{tab:threat_models}).
Consider $\mathcal{D}=\left\{\left(\boldsymbol{x}_i, y_i\right)\right\}_{i=1}^N$ as a clean training set, and $C: \mathcal{X} \rightarrow$ $\mathcal{Y}$ represents the functionality of the target neural network. For each sound $\boldsymbol{x}_i$ in $\mathcal{D}$, we have $\boldsymbol{x}_i \in \mathcal{X}=[0,1]^{C \times W \times H}$, and $y_i \in \mathcal{Y}=\{1, \ldots, J\}$, where $J$ is the number of label classes. To start an attack, backdoor adversaries must first poison (Figure \ref{fig:ML_data_poisoning}) the selected clean samples $\mathcal{D}_p$ via covert transformation $T(\cdot)$. The poisoned data are mixed with clean ones before training a backdoored model, which may be described as: $\mathcal{D}_t=\mathcal{D} \cup \mathcal{D}_p$, where $\mathcal{D}_p=\left(x_i^{\prime}, y_t\right) \mid x^{\prime}=T(x),\left(x_i, y_i\right) \in \mathcal{D}_p$. The deep neural network (DNN) is then optimized in the following:

$$
\min _{\boldsymbol{\Theta}} \sum_{i=1}^{N_b} \mathcal{L}\left(f\left(\boldsymbol{x}_i ; \boldsymbol{\Theta}\right), y_i\right)+\sum_{j=1}^{N_p} \mathcal{L}\left(f\left(\boldsymbol{x}_i^{\prime} ; \boldsymbol{\Theta}\right), y_t\right) .
$$ where $N_b=|\mathcal{D}|$ , $N_p=\left|\mathcal{D}_p\right|$.

\begin{table}[htbp]

\scriptsize   
\setlength{\tabcolsep}{1.0pt} 

	\centering
	\caption{Three Different Threat Models of Backdoor Attacks}
	\label{tab:threat_models}
	\begin{tabular}{@{}lccc@{}}
	\toprule
	Learning Task & Target Model & Training Dataset & Attacker Capability \\
	\midrule
	\tikz[baseline=(char.base)]{\node[circle,draw,fill=black,minimum size=1.5mm,inner sep=0pt] (char) {\textcolor{white}{$\mathcal{G}$}};} & : Full Knowledge & \tikz[baseline=(char.base)]{\node[circle,draw,fill=black,minimum size=1.5mm,inner sep=0pt] (char) {\textcolor{white}{$\mathcal{G}$}};} & : Full Knowledge \\
	\tikz[baseline=(char.base)]{\node[circle,draw,fill=black,minimum size=1.5mm,inner sep=0pt] (char) {\textcolor{white}{$\mathcal{G}$}};} & : Partial Knowledge & \tikz[baseline=(char.base)]{\node[circle,draw,fill=black,minimum size=1.5mm,inner sep=0pt,rotate=45] (char) {\textcolor{white}{$\mathcal{P}$}};} & : Partial Knowledge \\
	\tikz[baseline=(char.base)]{\node[circle,draw,fill=white,minimum size=1.5mm,inner sep=0pt] (char) {$\mathcal{N}$};} & : None Knowledge & \tikz[baseline=(char.base)]{\node[circle,draw,fill=white,minimum size=1.5mm,inner sep=0pt] (char) {$\mathcal{N}$};} & : None Knowledge \\
	\bottomrule
	\end{tabular}
\end{table}

\subsection{White-box Attack}

In a white-box attack, the attacker possesses complete knowledge of the learning task, the target model, and the training dataset, even if the attack only exploits a portion of it. This attack is the most favorable scenario for the attacker, a white-box attack can be denoted as:

\begin{equation}
	\text{Attacker Capability} =  \tikz[baseline=(char.base)]{\node[circle,draw,fill=black,minimum size=1.5mm,inner sep=0pt] (char) {\textcolor{white}{$\mathcal{G}$}};}
\end{equation}

\subsection{Grey-box Attack}

In a grey-box attack, the attacker knows the training objective, but has only incomplete information about the target model and the training dataset. A grey-box attack can be defined as follows:

\begin{equation}
	\text{Attacker Capability} = \tikz[baseline=(char.base)]{\node[circle,draw,fill=black,minimum size=1.5mm,inner sep=0pt] (char) {\textcolor{white}{$\mathcal{G}$}};}
  : \tikz[baseline=(char.base)]{\node[circle,draw,fill=black,minimum size=1.5mm,inner sep=0pt,rotate=45] (char) {\textcolor{white}{$\mathcal{P}$}};}
\end{equation}

\subsection{Black-box Attack}

In a black-box attack, neither the target model nor the training dataset, etc., is known to the attacker. This attack represents the most difficult situation for the attacker. A black-box attack can be represented as follows:

\begin{equation}
	\text{Attacker Capability} = \tikz[baseline=(char.base)]{\node[circle,draw,fill=white,minimum size=1.5mm,inner sep=0pt] (char) {$\mathcal{N}$};} : \tikz[baseline=(char.base)]{\node[circle,draw,fill=white,minimum size=1.5mm,inner sep=0pt] (char) {$\mathcal{N}$};}
\end{equation}

The adversary wishes to have the target model perform as predicted on benign data while working in the manner indicated by the adversary on poisoned samples. A formulation of the enemy's goal is: 

$$
\begin{aligned}
\min _{\mathcal{M}^*} \mathcal{L}\left(\mathcal{D}^b, \mathcal{D}^p, \mathcal{M}^*\right)= & \sum_{x_i \in \mathcal{D}^b} l\left(\mathcal{M}^*\left(x_i\right), y_i\right) \\
& +\sum_{x_j \in \mathcal{D}^p} l\left(\mathcal{M}^*\left(x_k^\star \circ  \varepsilon \right), y_t\right),
\end{aligned}
$$

where the benign and poisoned training datasets are denoted by $\mathcal{D}^b$ and $\mathcal{D}^p$, respectively. The function $l(\cdot, \cdot)$ represents the task-specific loss function. The integration of the backdoor trigger $(\varepsilon)$ into the training data is indicated by the symbol $\circ$.

\subsection{Poisoning Attack Capabilities}

We consider dataset $\mathcal{D} = \{ (x^{(i)}, y^{(i)}) \}_{i=1}^{N}$ or $\mathcal{D} = \{ (\mathcal{D}_{x}^{(i)}, \mathcal{D}_{y}^{(i)}) \}$. We denote the parameter initialization $\boldsymbol{\Theta}'$ and a training algorithm $\mathcal{M}$ as $\boldsymbol{\Theta} = \mathcal{M}(f, \boldsymbol{\Theta}', \mathcal{D}). $, i.e., given a model, initialization, and data, the training function $\mathcal{M}$ returns a trained parameterization $\boldsymbol{\Theta}$. Finally, we assume the loss function is computed element-wise from the dataset, $\mathcal{L}(f(x^{(i)}),y^{(i)})$.

\vspace{0.2cm}

\paragraph{Label Poisoning}
We describe the collection of possibly poisoned datasets as follows: Given a dataset $\mathcal{D}$, we represent the label poisoning \cite{sosnin2024certified} capabilities of an adversary as altering at most $l$ labels by magnitude at most $\zeta$ in a $\ell_q$ norm. 

\begin{align}
T_{h, \epsilon, p}^{l, \zeta, q}(\mathcal{D})  := \bigcup_{I \in \mathfrak{S}_h} \bigcup_{J \in \mathfrak{S}_l}  \{\mathcal{D}'\   \\ s.t.\ \forall i \in I,  ||\mathcal{D}'^{(i)}_x - \mathcal{D}^{(i)}_x ||_p \leq \epsilon \wedge \forall j \in J,  || \mathcal{D}'^{(j)}_y - \mathcal{D}^{(j)}_y ||_q \leq \zeta\}
\end{align}

Where $\mathfrak{S}_h$ is the set of all subsets.

\subsection{Poisoning Attack Goals}

\paragraph{Untargeted Poisoning} Untargeted poisoning aims to prevent training convergence. Given a test dataset of $\mathcal{C}$ examples, the adversary's objective as:

\begin{equation}\label{eq:untargeted}
    \max_{\mathcal{D}' \in T} \dfrac{1}{\mathcal{C}} \sum_{i=1}^{\mathcal{C}} \mathcal{L}\big( f^{\mathcal{M}(f, \boldsymbol{\Theta}', \mathcal{D}')}(x^{(i)}), y^{(i)} \big)
\end{equation}

\paragraph{Targeted Poisoning} The adversary ensures that the model's predictions fall outside a set of outputs $\mathscr{P}$. we formulate this as an optimization problem: 

\begin{equation}
\max_{\mathcal{D}' \in T} \dfrac{1}{\mathcal{C}} \sum_{i=1}^{\mathcal{C}} \mathbbm{1}\big( f^{\mathcal{M}(f, \boldsymbol{\Theta}', \mathcal{D}')}(x^{(i)}) \notin \mathscr{P} \big)
\label{eq:targeted}
\end{equation}

\paragraph{Backdoor Poisoning} 
The objective of the backdoor attack may be formulated as generating predictions outside a set $\mathscr{P}$ by assuming that the trigger manipulation(s) are confined to a set $\mathcal{F}(x)$. This is expressed as an optimization problem:

\begin{equation}\label{eq:backdoor}
\max_{\mathcal{D}' \in T} \dfrac{1}{\mathcal{C}} \sum_{i=1}^{\mathcal{C}} \mathbbm{1}\big( \exists k^\star \in \mathcal{F}(x^{(i)})\ s.t.\ f^{\mathcal{M}(f, \boldsymbol{\Theta}', \mathcal{D}')}(k^\star) \notin \mathscr{P} \big)
\end{equation}

\subsection{Backdoor attack Machine Learning}  

In the context of large language models (LLMs), adversary attack against DNNs focus on four \cite{li2024backdoorllm},\cite{li2024backdoor} main backdoor attack strategies: data poisoning (DP), weight poisoning (WP), hidden state (HA), and chain of thought (CA) attacks.

\begin{table}[H]
\centering
\caption{ backdoor attacks LLMs.}

\begin{adjustbox}{width=0.9\linewidth}
\begin{tabular}{c|ccc|c}
\toprule
\multirow{2}{*}{\begin{tabular}[c]{@{}c@{}}Backdoor\\ Attack\end{tabular}} & \multicolumn{3}{c|}{Access Requirement} & \multirow{2}{*}{\begin{tabular}[c]{@{}c@{}}Injection \\ Method\end{tabular}} \\ \cline{2-4}
 & Training Set & Model Weight & Internal Info &  \\ \hline
DP & \checkmark &  &  & supervised fine-tuning \\ 
WP &  & \checkmark & \checkmark & Model editing \\
HA &  & \checkmark & \checkmark & Activating the steering \\
CA &  &  & \checkmark &  Reasoning \\
\bottomrule
\end{tabular}
\end{adjustbox}
\label{tab:backdoor_summa}
\end{table}

\section{Adversarial Reinforcement learning in Finance via Bayesian Approach}
By conceptualizing according to the previous works studied in \cite{mengara2024trading},\cite{mengara2024last} we can deduce the following:

\subsection{Bayesian optimization application via Diffusion Model }

\begin{algorithm}[ht]
\caption{Diffusion Bayesian Optimization}
\label{alg:Diffusion_bayesian_optimization}
\KwData{T, $\theta$, $\alpha$, $\beta$, $\sigma$}
\KwResult{Model parameters and trace.}
Initialize $x_T$\;
\For{$t \leftarrow T - 1$ \textbf{downto} $0$}{
    \If{$t > 1$}{
        $z \gets$ Noise\_dist$(0)$\;
    }
    Else{
        $z \gets 0$\;
    }
    $transport\_component \gets$ Optimal\_transport$(x_T, t, \theta, \beta, \sigma)$\;
    $x_{t-1} \gets$ Normal$(f'x_t', \mu = drift\_function(x_T, t, \theta, \beta, \sigma) + transport\_component + \sigma[t] \cdot z, \sigma = 1)$\;
    $x_T \gets x_{t-1}$\;
}
\end{algorithm}

\subsection{Bayesian optimization application via Limit Order Markets }

Consider an agent at time $t$ who wishes to maximise her expected utility by allocating her wealth in a risk-free bank account or a risky asset. Let us define the following processes:
 $B=\left(B_t\right)_{0 \leq t \leq T}$ is the risk-free bank account and satisfies:

$$
\mathrm{d}B_t=r B_t\mathrm{~d}t; 
$$

$$
\mathrm{d} S_t=(\mu-r) S_t \mathrm{~d} t+\sigma S_t \mathrm{~d} W_t, S_0=s
$$

where, $W=\left(W_t\right)_{0 \leq t \leq T}$ is a Brownian motion,
$S=\left(S_t\right)_{0 \leq t \leq T}$ is the discounted risky price process.

$\pi=\left(\pi_t\right)_{0 \leq t \leq T}$ is a self-financing strategy, which indicates the amount of money allocated in the risky asset at time $t$.

$X^\pi=\left(X_t^\pi\right)_{0 \leq t \leq T}$ is the agent's discounted wealth given the strategy $\pi$, and satisfies the following stochastic differential equation:

$$
\mathrm{d} X_t^\pi=\left(\pi_t(\mu-r)+r X_t^\pi\right) \mathrm{d} t+\pi_t \sigma \mathrm{~d} W_t, X_0^\pi=x .
$$

The maximisation problem is formulated as follows:

$$
H^{\pi, t}(s, x)=\sup _{=\in \mathcal{A}_{0, T}} \mathbb{E}_{s, x}\left[U\left(X_T^\pi\right)\right]
$$

where $U(x)$ is the agent's utility function, $\mathcal{A}_{t, T}$ is the set of all admissible strategies (Table \ref{fig:Order_Book}), corresponding to all $\mathcal{F}$-predictable self-financing strategies such that $\int_t^T \pi_s^2 \mathrm{~d} s<\infty$. 

\begin{table}[H]
\centering
\caption{Comparison of Market and Limit Orders.}
\begin{adjustbox}{width=1.0\linewidth}
\begin{tabular}{|c|c|c|c|c|c}
\hline  Feature & Market Orders & Limit Orders  \\
\hline  Execution Speed & Immediate & May take time \\
\hline  Price Control & None & Yes \\
\hline   Risk of Non-Execution & No & Yes \\
\hline  Best For & Urgent trades & Price-sensitive trades \\
\hline  Impact of Volatility & High risk of slippage & Protects against adverse price movements \\
\hline
\end{tabular}
\end{adjustbox}
\label{fig:Order_Book}
\end{table}

Let's the optimal liquidation \footnote{\href{https://www.quantstart.com/articles/high-frequency-trading-iii-optimal-execution/}{High Frequency Trading}} \footnote{\href{https://quant.stackexchange.com/questions/46125/what-mathematical-theory-is-required-for-high-frequency-trading}{Mathematical theory}} \footnote{\href{https://github.com/quantrocket-codeload}{QuantRocket}} \footnote{\href{https://github.com/lcsrodriguez/optimalHFT/blob/main/main.ipynb}{High Frequency}} speed. Let $\mathcal{A}$ be the set of all predictable non-negative bounded processes. Our set of admissible strategies that is, the liquidation (algorithm \ref{alg:simulate_execution}) speed $\nu$ will have to be picked from $\mathcal{A}$. 

Suppose that we want to liquidate \footnote{\href{https://www.finra.org/investors/investing/investment-products/stocks/order-types}{Order Types: Market, Limit, and Stop Orders}} \cite{wei2019model} a portfolio of $\mathcal{P}$ shares by a terminal time $T$. Then, our objective will be to minimize:

$$
\mathbb{E}_{t, S, q}\left[\int_t^T S_u^\nu \nu_u d u+\left(\mathcal{P}-Q_T^\nu\right) S_T+\alpha\left(\mathcal{P}-Q_T^\nu\right)^2\right]
$$

over all possible strategies $\nu \in \mathcal{A}$, and where $\alpha>0$. That is, we would like to find the value function. 

$$
H(t, S, q)=\inf _{\nu \in \mathcal{A}} \mathbb{E}_{t, S, q}\left[\int_t^T \hat{S}_u^\nu \nu_u d u+\left(\mathcal{P}-Q_T^\nu\right) S_T+\alpha\left(\mathcal{P}-Q_T^\nu\right)^2\right]
$$

\begin{mdframed}[leftmargin=0pt, rightmargin=0pt, innerleftmargin=10pt, innerrightmargin=10pt, skipbelow=0pt]
\faLightbulbO~\textit{The first term represents the amount of cash we obtain by following some strategy $\nu$. The second term, on the other hand, indicates that the trader must execute all the shares that were not liquidated at time $T$. Finally, the third term is a terminal penalty, where we penalise not liquidating all shares by time $T$.}

\end{mdframed}

By Hamilton-Jacobi-Bellman equation to deduce that the value function $H$ must satsify the following Partial Differential Equation:

$$
\left\{\begin{array}{l}
\partial_t H+\frac{1}{2} \sigma^2 \partial_{S S} H+\inf _{\nu \in \mathcal{A}}\left\{(S+k \nu) \nu-\nu \partial_q H\right\} \\
H(T, S, q)=(\mathcal{P}-q) S+\alpha(\mathcal{P}-q)^2
\end{array}\right.
$$

The optimal liquidation speed is given by,

$$
\nu_t^*=\frac{\mathcal{P}}{T+\frac{k}{\alpha}}
$$

\begin{algorithm}[ht]
\SetAlgoLined
\KwData{num\_steps = 10000}
\KwResult{Simulated stock price $S$, market order quantity $Q_{mkt}$, limit order quantity $Q_{lim}$, and total liquidated shares $Q_{total}$}

Initialize arrays $S$, $Q_{mkt}$, $Q_{lim}$, and $Q_{total}$ of size num\_steps\;
$S[0] = 100$\; \# Initial stock price\;
$Q_{mkt}[0] = \mathcal{P} * \mu / (\mu + \theta)$\;
$Q_{lim}[0] = \mathcal{P} * \mu / (\mu + \beta)$\;

\For{i from 1 to num\_steps - 1}{
    $\nu^{*}_{mkt} = \frac{\mathcal{P}}{T + k / \alpha}$\;
    $\nu^{*}_{lim} = \nu^{*}_{mkt} * \frac{\beta}{\mu}$\;

    \# Market order execution\;
    $dS_{mkt} = \sigma * \sqrt{dt} * \mathcal{N}(0,1) + \theta * \nu^{*}_{mkt} * dt$\;
    $dQ_{mkt} = -\nu^{*}_{mkt} * dt$\;

    $S[i] = S[i-1] + dS_{mkt}$\;
    $Q_{mkt}[i] = \max(Q_{mkt}[i-1] + dQ_{mkt}, 0)$\;

    \# Limit order execution\;
    $\phi = \Phi(\gamma * \sqrt{dt})$\;
    $dQ_{lim} = -\nu^{*}_{lim} * dt * \phi$\;

    $Q_{lim}[i] = \max(Q_{lim}[i-1] + dQ_{lim}, 0)$\;

    \# Update total liquidated shares\;
    $Q_{total}[i] = Q_{mkt}[i] + Q_{lim}[i]$\
}
\caption{Simulate market and limit order executions}
\label{alg:simulate_execution}
\end{algorithm}

\begin{algorithm}[ht]
\SetAlgoLined
\KwData{$prices$, $bid\_ask\_spreads$, $liquidity\_factor$}
\KwResult{Cumulative profit, cumulative slippage}

$n_{trades} = \text{len}(prices)$\;
$slippage = []$\;
$profits = []$\;

\For{i from 1 to $n_{trades} - 1$}{
    $trade\_price = prices[i] + \mathcal{U}(-bid\_ask\_spreads[i], bid\_ask\_spreads[i])$\;
    
    $trade\_slippage = |trade\_price - prices[i]|$\;
    $slippage.append(trade\_slippage)$\;

    $profit = prices[i] - prices[i-1] - trade\_slippage$\;
    $profits.append(profit)$\;

    $cumulative\_profit = \sum profits$\;
    $cumulative\_slippage = \sum slippage$\;
}

\caption{Simulate high frequency trading}
\label{alg:simulate_high_frequency_trading}
\end{algorithm}

Base approach for modeling High Frequency Trading (algorithm \ref{alg:simulate_high_frequency_trading})\cite{zaharudin2022high},\cite{miller2016high}, \cite{gomber2015high}, \cite{brogaard2014high},\cite{o2015high}, \cite{hagstromer2013diversity}, \cite{goettler2005equilibrium}, \cite{tsantekidis2023modeling}, \cite{sun2022market}, \cite{harris2002trading}. HFTs together with a discrete price model derived from discrete portfolio execution theories is: A time period $T$ into $N$ even short interval of length $T=T / N, S_m$ is the security price at time $t=n \tau$.

$$
S_t=S_{t-1}+\sigma \pi^{1 / 2} \xi_i-\pi g\left(\frac{h_{\mathrm{t}} t_2}{\tau}\right)+\delta \tau_t
$$

For $t=1, \ldots, N, \sigma$ repersents the volatility, and $\xi_i \sim N(0 ; 1)$. $h_{\text {}}$ represents the net sale volumes of all HFT , $g(v)$ is the price impact function.

\subsection{Bayesian optimization application via stochastic volatility jump}

\begin{figure}[H]
\centering
\includegraphics[width=0.42\textwidth]{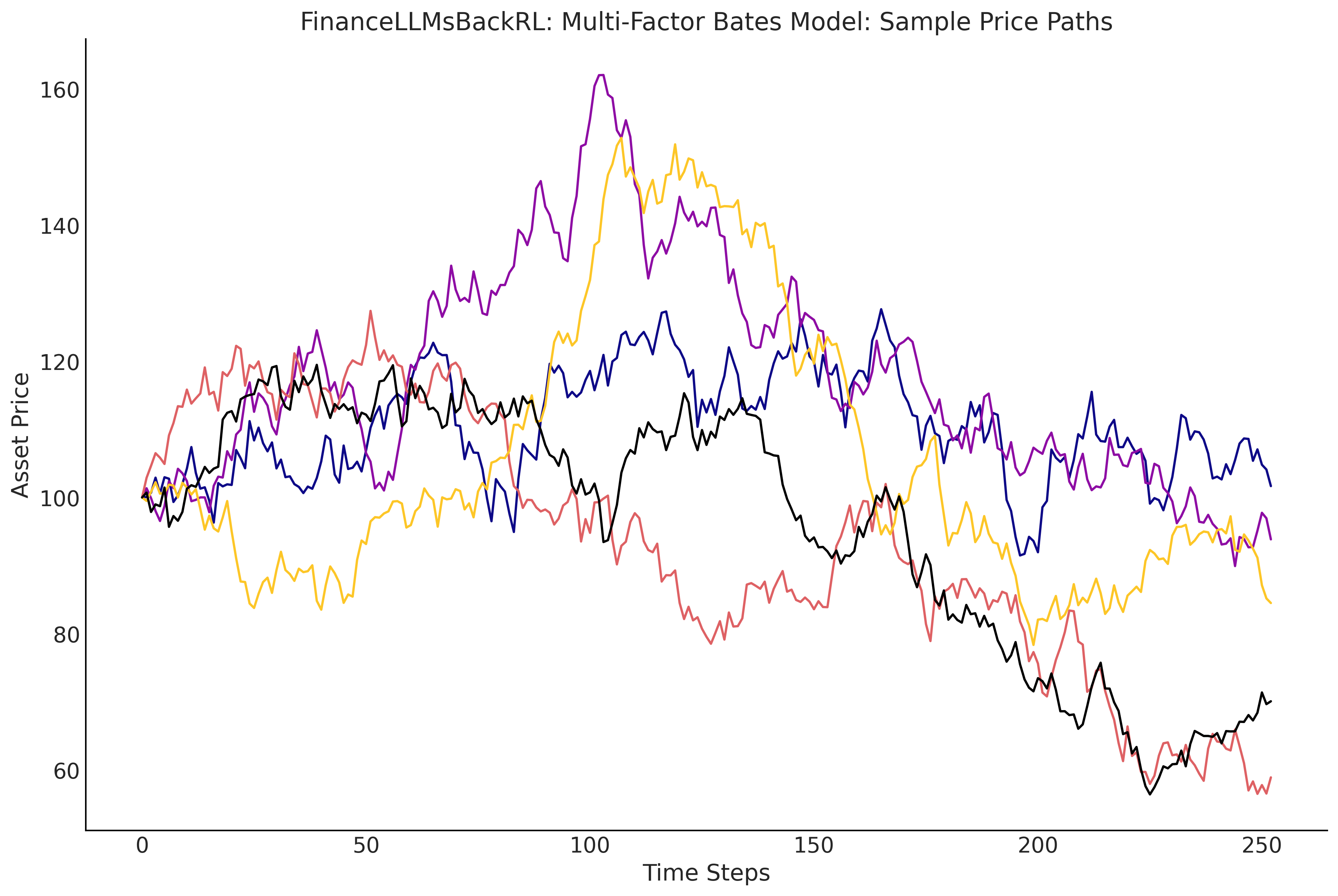}
\caption{stochastic volatility jump.}
\label{fig:Bates_model}
\end{figure}

The Bates model (stochastic volatility jump)\cite{bates1996jumps},\cite{deng2020option},\cite{huynh2022subgraph},\cite{nunes2011analytic},\cite{soleymani2019pricing} (Figure \ref{fig:Bates_model}) is defined by two coupled stochastic differential equations:

\vspace{0.2cm}

 $ds(t)=(r-\lambda k) s(t) dt+\int v(t) s(t) dw_1(t)+J(t) s(t) dN(t)$ 

 $dv(\mathrm{t})=\kappa(\theta-v(\mathrm{t})) \mathrm{dt}+\sigma \sqrt{v}(\mathrm{t}) dw_2(t) $

\vspace{0.2cm}

Where: $\mathrm{s}(\mathrm{t})$ is the asset price,
$v(t)$ is the variance, $r$ is the risk-free rate, $\lambda$ is the jump intensity, k is the expected relative jump size, $\kappa, \theta, \sigma$ are volatility parameters,  $w_1, w_2$ are Wiener processes with correlation $\rho$, $J(t)$ jump size, $N(t)$ is a Poisson process.

\subsection{Bayesian optimization application via CIR (Cox-Ingersoll-Ross) Model}

$$
d r(t)=\kappa(\theta-r(t)) d t+\sigma \sqrt{r(t)} d W(t)
$$

The Cox-Ingersoll-Ross model \cite{bernaschi2007empirical},\cite{overbeck1997estimation} is to guarantee a non-negative short rate \cite{cozma2020strong} \footnote{\href{https://quant.stackexchange.com/questions/8114/monte-carlo-simulating-cox-ingersoll-ross-process}{Monte Carlo simulating CIR}} \footnote{\href{https://math.stackexchange.com/questions/3328585/the-cox-ingersoll-ross-model-1985}{CIR :mathematical element}} \footnote{\href{https://quant.stackexchange.com/questions/44475/feller-condition-cox-ingersoll-ross-source}{CIR: Feller Condition}} \footnote{\href{https://github.com/dpicone1/Vasicek_CIR_HoLee_HullWhite_Models_Python/blob/master/HoLee_Model_in_Python.ipynb}{HoLee Model}} model stays strictly positive if we have,

$$
2 \kappa \theta>\sigma^2
$$

$$
g(t)=\frac{4 \kappa e^{-\kappa t}}{\sigma^2\left(1-e^{-\kappa t}\right)}, \lambda=r_0 g(t), d=4 \kappa \theta / \sigma^2
$$

$$
\lim _{t \rightarrow \infty} E(r(t))=\theta, \lim _{t \rightarrow \infty} \operatorname{Var}(r(t))=\frac{\theta a^2}{2 \kappa}
$$

\begin{algorithm}[ht]
\SetAlgoLined
\KwData{a, b, $\sigma$, r, T, N}
\KwResult{Time series of interest rate r}

$dt = \frac{T}{N}$\;
$t = \left(0, T, \frac{N}{100}\right)$\; %
$r = zeros(int(N) + 1)$\; %

$r[0] = r$\;

\For{i from 1 to int(N)}{
    $r[i] = r[i-1] + a(b-r[i-1])dt + \sigma\sqrt{r[i-1]}dt\mathcal{N}(0, 1)$\;
}

\caption{CIR}
\label{alg:cir_drift}
\end{algorithm}

\begin{figure}[H] 
\centering
\includegraphics[scale=0.38]{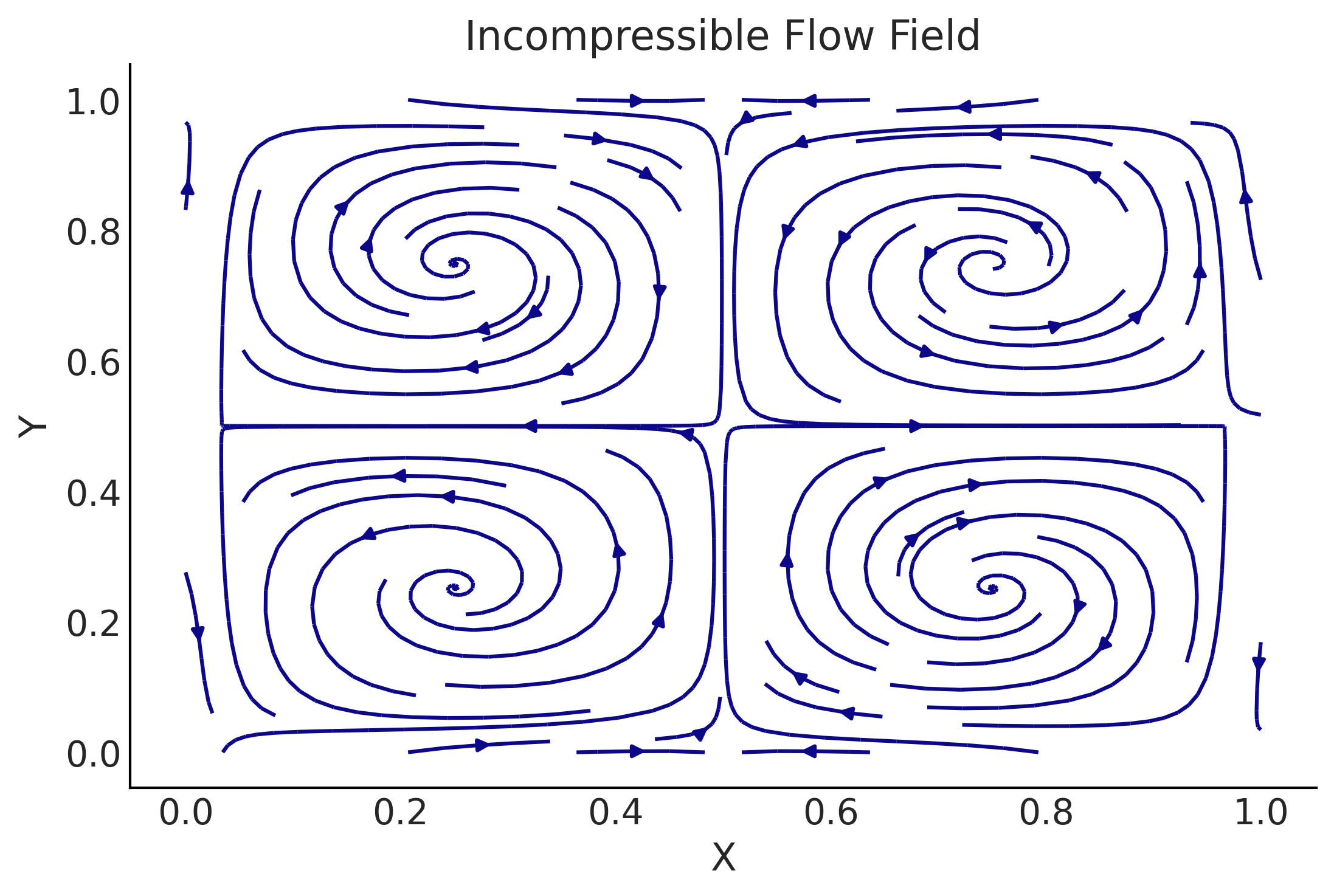}
\caption{FinanceLLMsBackRL: Incompressible flow. }
\label{fig:incompressible_flow}
\end{figure}

\subsection{Bayesian optimization application via Navier-Stokes equations}

The Euler and Navier-Stokes equations \cite{scheffer1993inviscid},\cite{lin1998new},\cite{ladyzhenskaya1969mathematical} \cite{constantin2001some},\cite{bertozzi2002vorticity} \cite{caffarelli1982partial},\cite{bradshaw2018self} \cite{lemarie2018navier},\cite{temam2024navier},\cite{tao2019quantitative},\cite{tao2019searching},\cite{tao2016finite},\cite{barker2024concentration} describe the motion of a fluid in $\mathbb{R}^n$ ($n=2$ or $3$).

These equations \cite{albritton2022non},\cite{guillod2023numerical},\cite{chen2022stable} are  solved for an unknown velocity vector $u(x, t)=\left(u_i(x, t)\right)_{1 \leq i \leq n} \in \mathbb{R}^n$ and pressure $p(x, t) \in \mathbb{R}$, defined for position $x \in \mathbb{R}^n$ and time $t \geq 0$, to incompressible (Figure \ref{fig:incompressible_flow}) \cite{wang2024recent} fluids filling all of $\mathbb{R}^n$. The Navier-Stokes equations are then given by,

$$
\begin{aligned}
& \frac{\partial}{\partial t} u_i+\sum_{j=1}^n u_j \frac{\partial u_i}{\partial x_j}=\nu \Delta u_i-\frac{\partial p}{\partial x_i}+f_i(x, t) \quad\left(x \in \mathbb{R}^n, t \geq 0\right), \\
& \operatorname{div} u=\sum_{i=1}^n \frac{\partial u_i}{\partial x_i}=0 \quad\left(x \in \mathbb{R}^n, t \geq 0\right)
\end{aligned}
$$

$$
u(x, 0)=u^{\circ}(x) \quad\left(x \in \mathbb{R}^n\right)
$$

Where, $u^{\circ}(x)$ is a given, $C^{\infty}$ divergence-free vector field on $\mathbb{R}^n, f_i(x, t)$ are the components of a given, externally applied force, $\nu$ is a positive coefficient (the viscosity), and $\Delta=\sum_{i=1}^n \frac{\partial^2}{\partial x_i^2}$ is the Laplacian in the space variables.

$$
\left|\partial_x^\Gamma u^{\circ}(x)\right| \leq C_{\Gamma K}(1+|x|)^{-K} \quad \text { on } \mathbb{R}^n, \text { for any } \Gamma \text { and } K
$$

 $\left|\partial_x^\Gamma \partial_t^\eta f(x, t)\right| \leq C_{a \eta K}(1+|x|+t)^{-K} \quad$ on $\mathbb{R}^n \times[0, \infty)$, for any $\Gamma, \eta, K$.

$$
p, u \in C^{\infty}\left(\mathbb{R}^n \times[0, \infty)\right)
$$

$$
\int_{\mathbb{R}^n}|u(x, t)|^2 d x<C \quad \text {   for all   }  t \geq 0 \quad 
$$

$$
u^{\circ}\left(x+e_j\right)=u^{\circ}(x), \quad f\left(x+e_j, t\right)=f(x, t) \quad \text { for  }  1 \leq j \leq n
$$

 $\left|\partial_x^\Gamma \partial_t^\eta f(x, t)\right| \leq C_{a \eta K}(1+|t|)^{-K} \quad$ on $\mathbb{R}^3 \times[0, \infty)$, for any $\Gamma, \eta, K$.

$$
u(x, t)=u\left(x+e_j, t\right)
$$

on $\mathbb{R}^3 \times[0, \infty)$ for $1 \leq j \leq n$

$$
p, u \in C^{\infty}\left(\mathbb{R}^n \times[0, \infty)\right)
$$

 $p\left(x+e_j, t\right)=p(x, t)$ , 
 
$$
\begin{aligned}
& -\iint_{\mathbb{R}^3 \times \mathbb{R}} u \cdot \frac{\partial \theta}{\partial t} \mathrm{~d} x \mathrm{~d} t-\sum_{i j} \iint_{\mathbb{R}^3 \times \mathbb{R}} u_i u_j \frac{\partial \theta_i}{\partial x_j} \mathrm{~d} x \mathrm{~d} t \\
& = \hspace{-1cm}\nu \iint_{\mathbb{R}^3 \times \mathbb{R}} u \cdot \Delta \theta \mathrm{~d} x \mathrm{~d} t+\iint_{\mathbb{R}^3 \times \mathbb{R}} f \cdot \theta \mathrm{~d} x \mathrm{~d} t+\iint_{\mathbb{R}^3 \times \mathbb{R}} p \cdot(\operatorname{div} \theta) \mathrm{d} x \mathrm{~d} t
\end{aligned}
$$

\begin{algorithm}[ht]
\SetAlgoLined
\KwData{$Lx$, $Ly$, $Lz$, $Nx$, $Ny$, $Nz$}
\KwResult{velocities $u$, $v$, $w$, and grid spacings $dx$, $dy$, $dz$}

$dx = \frac{Lx}{Nx - 1}; dy = \frac{Ly}{Ny - 1}; dz = \frac{Lz}{Nz - 1}$\;

\vspace{0.2cm}

$r = \sqrt{x^2 + y^2 + z^2}$\;
$u[:] = \sin(r) * \exp(-r)$\;
$v[:] = \cos(r) * \exp(-r)$\;
$w[:] = \sin(r) * \exp(-r)$\;

\vspace{0.2cm} 

 velocity profiles\;
$u[:, :, :] = 1.0 * (\sin(\pi * \frac{k}{Nx}) * \cos(\pi * \frac{j}{Ny}) *
                     \exp(-(k/Nz)^2))$\;
$v[:, :, :] = 0.5 * (\sin(\pi * \frac{j}{Ny}) * \cos(\pi * \frac{i}{Nx}) *
                     \exp(-(k/Nz)^2))$\;
$w[:, :, :] = 0.25 * (\sin(\pi * \frac{k}{Nz}) * \cos(\pi * \frac{i}{Nx}) *
                      \exp(-(j/Ny)^2))$\;

\Return{$u$, $v$, $w$, $dx$, $dy$, $dz$} 

\caption{Initialize system}
\label{alg:initialize_system}
\end{algorithm}

\begin{algorithm}
\SetAlgoLined
\KwData{$u$, $v$, $w$, $dx$, $dy$, $dz$, $dt$, $\rho$, $\mu$}
\KwResult{Updated velocities $u_{new}$, $v_{new}$, $w_{new}$}.
 
\vspace{0.2cm}
\text{Compute Reynolds number} $Re$\;
$du_dx = \frac{\text{roll}(u, -1, axis=1) - u}{dx}$\;
$dv_dy = \frac{\text{roll}(v, -1, axis=0) - v}{dy}$\;
$dw_dz = \frac{\text{roll}(w, -1, axis=2) - w}{dz}$\;

\vspace{0.2cm}
\text{Compute pressure gradient}  $grad_{p}$\;

$u_{new} = u - dt * (grad_p + \sum \tau_{ij})$\;
$v_{new} = v - dt * (grad_p + \sum \tau_{ij})$\;
$w_{new} = w - dt * (grad_p + \sum \tau_{ij})$\;

\Return{$u\_new$, $v\_new$, $w\_new$} 

\caption{Navier-Stokes equations}
\label{alg:navier_stokes}
\end{algorithm}

\begin{algorithm}[ht]
\SetAlgoLined
\KwData{$u$, $v$, $w$, $dx$, $dy$, $dz$, $dt$, $\rho$, $\mu$}
\KwResult{Drag coefficient $cd$}

Define functions $force_{balance}$, $dudx$, $dudy$, $dwdz$\;
Compute total forces in $x$, $y$, $z$ directions\;

$drag\_force = \sqrt{\text{total\_force}_x^2 + \text{total\_force}_y^2 + \text{total\_force}_z^2}$\;
$drag\_area = 1.0$\; 
$velocity = \sqrt{u^2 + v^2 + w^2}.mean()$\;
$cd = drag\_force / (0.5 * density * velocity^2 * drag\_area)$\;
\Return{$u\_new$, $v\_new$, $w\_new$} 
\caption{Compute drag coefficient}
\label{alg:compute_drag_coefficient}
\end{algorithm}

\subsection{Bayesian optimization application via Reinforcement learning }

\begin{algorithm}[ht]
\SetAlgoLined

\KwData{Sampling rate, Imperceptibility}
\KwResult{Trigger object}

\SetKwFunction{initialize}{__init__}
\SetKwFunction{generateDynamicTrigger}{generate_dynamic_trigger}

\SetKwBlock{GenerateDynamicTrigger}{generate dynamic trigger}{}
\GenerateDynamicTrigger{}{
    $\text{state} \leftarrow \text{None}$\;
    \While{$\text{state} < \text{sampling\_rate}$}{
        $\text{action} \leftarrow \mathcal{U}(0, 1)$\;
        $\text{end\_state} \leftarrow \text{state} + 1$\;
        $\text{reward} \leftarrow \text{calculate\_reward}(\text{state}, \text{end\_state})$\;
        $\text{q\_table}[\text{state}, \text{action}] \leftarrow \text{q\_table}[\text{state}, \text{action}] + \text{learning\_rate} \cdot (\text{reward} + \text{discount\_factor} \cdot \max\limits_{a} \text{q\_table}[\text{end\_state}, a])$\;
        $\text{state} \leftarrow \text{end\_state}$\;
    }
    \Return{$\text{Trigger}(\text{sampling\_rate}=\text{sampling\_rate}, \text{imperceptibility}=\text{imperceptibility})$\;}
}

\caption{Reinforcement Learning Trigger}
\label{alg:reinforcement_learning_trigger}

\end{algorithm}

In the infinite horizon \cite{sun2023reinforcement},\cite{pippas2024evolution},\cite{zhang2019deep} setting \footnote{\href{https://gist.github.com/sebjai?direction=desc&sort=updated}{RL Finance}} \footnote{\href{https://qlib.readthedocs.io/en/latest/component/rl/overall.html}{RL Quantitative finance}}, an MDP (Markov Decision Process) is said to be linear with a feature map $\phi: \mathcal{S}\times \mathcal{A} \rightarrow \mathbb{R}^d$, if  there exist $d$ unknown measures $\mu = (\mu^{(1)},\cdots,\mu^{(d)})$ over $\mathcal{S}$ and an unknown vector $\theta \in \mathbb{R}^d$ such that for any $(s,a)\in \mathcal{S}\times \mathcal{A}$,

\begin{eqnarray}\label{eq:linearMDP_infinite}
P(\cdot|s,a) = \langle\, \phi(s,a),\mu(\cdot)\,\rangle,\quad r(s,a) = \langle\, \phi(s,a),\theta\,\rangle.
\end{eqnarray}

In the finite horizon \cite{liu2024dynamic} \cite{liu2022finrl},\cite{liu2018practical},\cite{halperin2017qlbs}, setting, an  MDP is said to be linear with a feature map $\phi: \mathcal{S}\times \mathcal{A} \rightarrow \mathbb{R}^d$, if for any $0\leq t \leq T$, there exist $d$ unknown measures $\mu_t = (\mu_t^{(1)},\cdots,\mu_t^{(d)})$ over $\mathcal{S}$ and an unknown vector $\theta_t \in \mathbb{R}^d$ such that for any $(s,a)\in \mathcal{S}\times \mathcal{A}$,

\begin{eqnarray}\label{eq:linearMDP_finite}
P_t(\cdot|s,a) = \langle\, \phi(s,a),\mu_t(\cdot)\,\rangle,\quad r_t(s,a) = \langle\, \phi(s,a),\theta_t\,\rangle,
\end{eqnarray}

$\|\phi(s,a)\|\leq 1$ for all $(s,a)\in \mathcal{S}\times \mathcal{A}$.

\begin{eqnarray}\label{eq:lfa_infinite}
Q(s,a) = \langle\, \psi(s,a), \omega \,\rangle, \quad v(s) = \langle\, \xi(s), \eta \,\rangle
\end{eqnarray}

\begin{eqnarray}\label{eq:lfa_finite}
Q_t(s,a) = \langle\, \psi(s,a), \omega_t \,\rangle, \quad v_t(s) = \langle\, \xi(s), \eta_t \,\rangle, \forall\, 0\leq t \leq T
\end{eqnarray}

$\psi: \mathcal{S}\times \mathcal{A}\rightarrow \mathbb{R}^d$ and $\xi: \mathcal{S}\rightarrow \mathbb{R}^d$ are known feature mappings and $\omega$, $\omega_t$ $\eta$, and $\eta_t$ are unknown vectors.

A victim agent  backdoor attack follow the \cite{yu2024spatiotemporal},\cite{chen2022marnet},\cite{kiourti2019trojdrl},\cite{cui2024badrl},\cite{bharti2022provable},\cite{guo2023policycleanse},\cite{lei2023new},\cite{ilahi2021challenges} following policy:

\begin{equation}
\pi_\text{Poison}(s)=\left\{
\begin{aligned}
\pi_\text{fail}(s) &  \text{, if }  \text{triggered} \\
\pi_\text{win}(s) &  \text{, if  }  \text{otherwise} \\
\end{aligned}
\right.
\label{eq:mixed}
\end{equation}

$$
\sum_{t=0}^\infty\gamma^t(c-R_1(s^{(t)}, a_1^{(t)}, s^{(t+1)})).
$$

\begin{figure}
\centering
\includegraphics[width=0.44\textwidth]{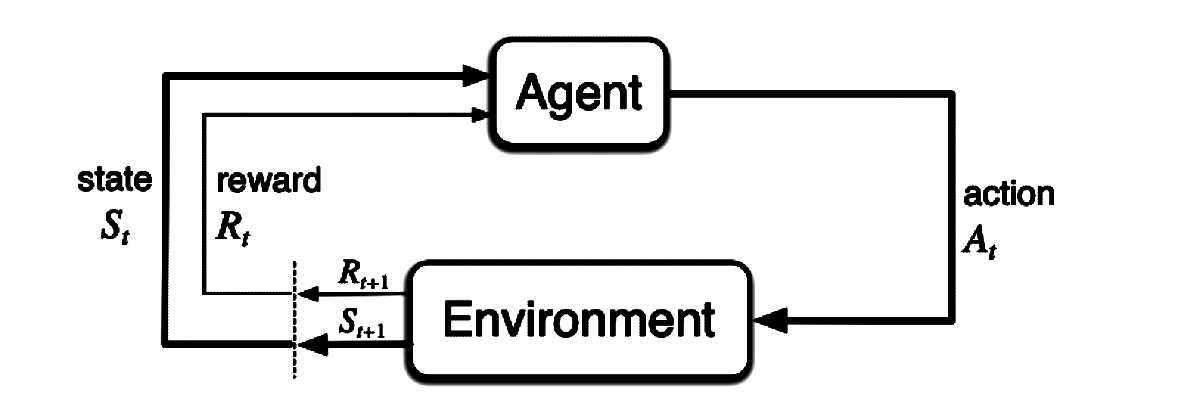}
\caption{RL: Environemment.}
\label{fig:Environemment_RL}
\end{figure}

\begin{figure}
\centering
\includegraphics[width=0.39\textwidth]{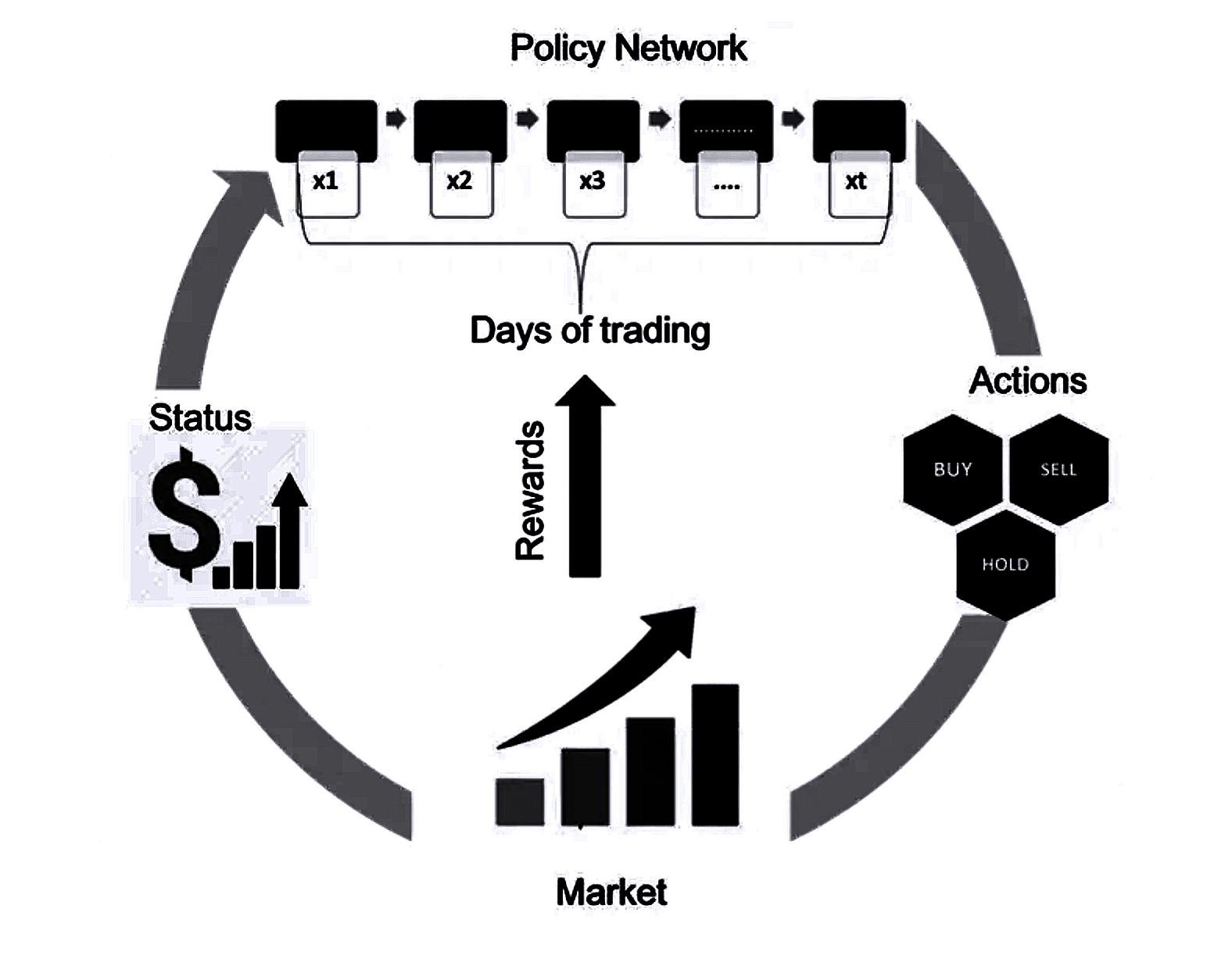}
\caption{RL: Data Poisoning.}
\label{fig:attacker's_trading_RL}
\end{figure}

We define the expected reward (Figure \ref{fig:attacker's_trading_RL}) for a policy $\pi$ \cite{haarnoja2018soft},  used in an environment (Figure \ref{fig:Environemment_RL}) as $\mathcal{E}$ by,

\begin{equation}
    R(\pi,\mathcal{E})=\mathbb{E}_{T\sim p(T|\pi,\mathcal{E})}\left[\sum_t r(s_t,a_t)\right]
\end{equation}

In a clean environment $\mathcal{E}$, the attacker wants to obtain a policy $\widetilde{\pi}$ that yields an expected reward comparable to that of the conventional model,

\begin{equation}
    \left|R(\pi^*,\mathcal{E})-R(\widetilde{\pi},\mathcal{E})\right|<\epsilon_1
\end{equation}

When the trigger is present in the environment, we refer to this as the poisoned environment $\widetilde{\mathcal{E}}$.

\begin{equation}
\max\ \big(R(\pi^*,\mathcal{E})-R(\widetilde{\pi},{\widetilde{\mathcal{E}}})\big)
\end{equation}

\begin{equation}
    \left|R(\pi^*,\mathcal{E})-R(\pi^*,\widetilde{\mathcal{E}})\right|<\epsilon_2
\end{equation}


\begin{figure}
\centering
\includegraphics[scale=0.30]{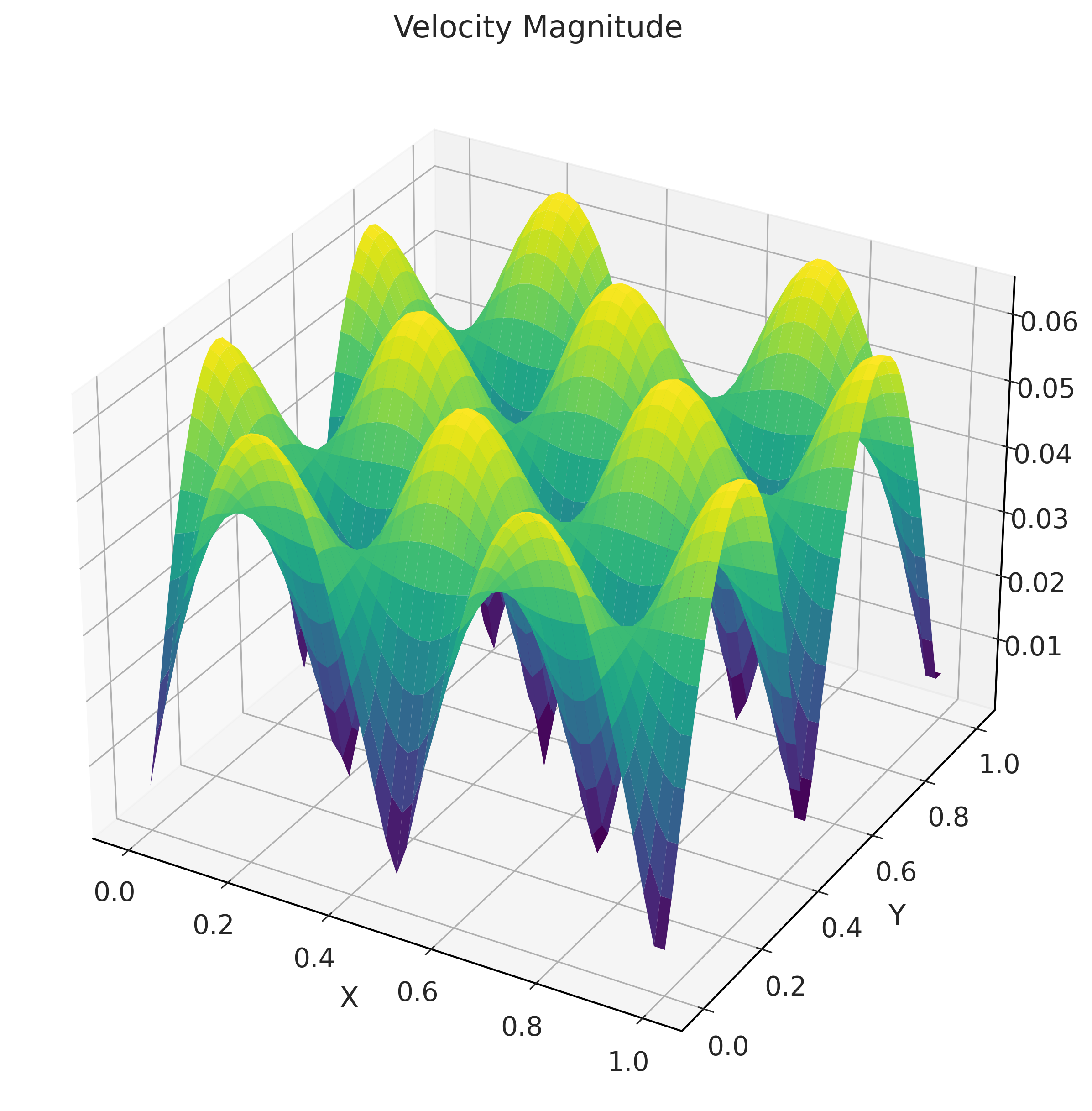} 
\caption{FinanceLLMsBackRL: Velocity Magnitude. }
\label{fig:velocity_magnitude}
\end{figure}

\section{FinanceLLMsBackRL: Bayesian Computational Modeling (LLM-RL) by Attack Scenario} 

In this study , we are inspired by the mathematical models of portfolios \cite{day2024portfolio},\cite{liang2018adversarial},\cite{abergel2013mathematical},\cite{zhang2024optimizing},\cite{pricope2021deep} investment \footnote{\href{https://personal.ntu.edu.sg/nprivault/indext.html}{Dr. Nicolas Privault}} \footnote{\href{https://martin-haugh.github.io/teaching/}{Dr. Martin Haugh}} model; High-Frequency Trading \cite{qin2024earnhft},\cite{briola2021deep} \footnote{\href{https://en.wikipedia.org/wiki/High-frequency_trading}{High Frequency Trading }} \footnote{\href{https://hftbacktest.readthedocs.io/en/py-v2.1.0/index.html}{High-Frequency Trading Backtesting Tool}} \cite{jacob2016rock},\cite{crawford2018automatic},\cite{almgren2001optimal};\cite{creswell2010speedy},\cite{kumar2023deep};\cite{gueant2017optimal}; Navier-Stokes equations existence and smoothing \cite{jormakka2008solutions},\cite{lai2024self}.“FinanceLLMsBackRL” is a technique that implements a poisoning attack with a ”clean-label backdoor“.

\vspace{0.3cm}

We propose a new adversarial framework for the design of selected sample triggers by “FinanceLLMBackRL” backdoor poisoning attacks on audio data applied to transformers via LLMs (text generators), showing that backdoor attacks applied only to audio data can transfer via other critical applications directly incorporating large language models in their operation chains without any assumptions. Our approach focuses on developing new and more advanced financial simulation methods using state-of-the-art Bayesian optimization methods with diffusion models \footnote{\href{https://huggingface.co/blog/Esmail-AGumaan/diffusion-models}{Diffusion Models HuggingFace}} (drift functions, including Bayesian diffusion optimization).

\vspace{0.3cm}
  
\begin{figure}[H] 
\centering
\includegraphics[scale=0.20]{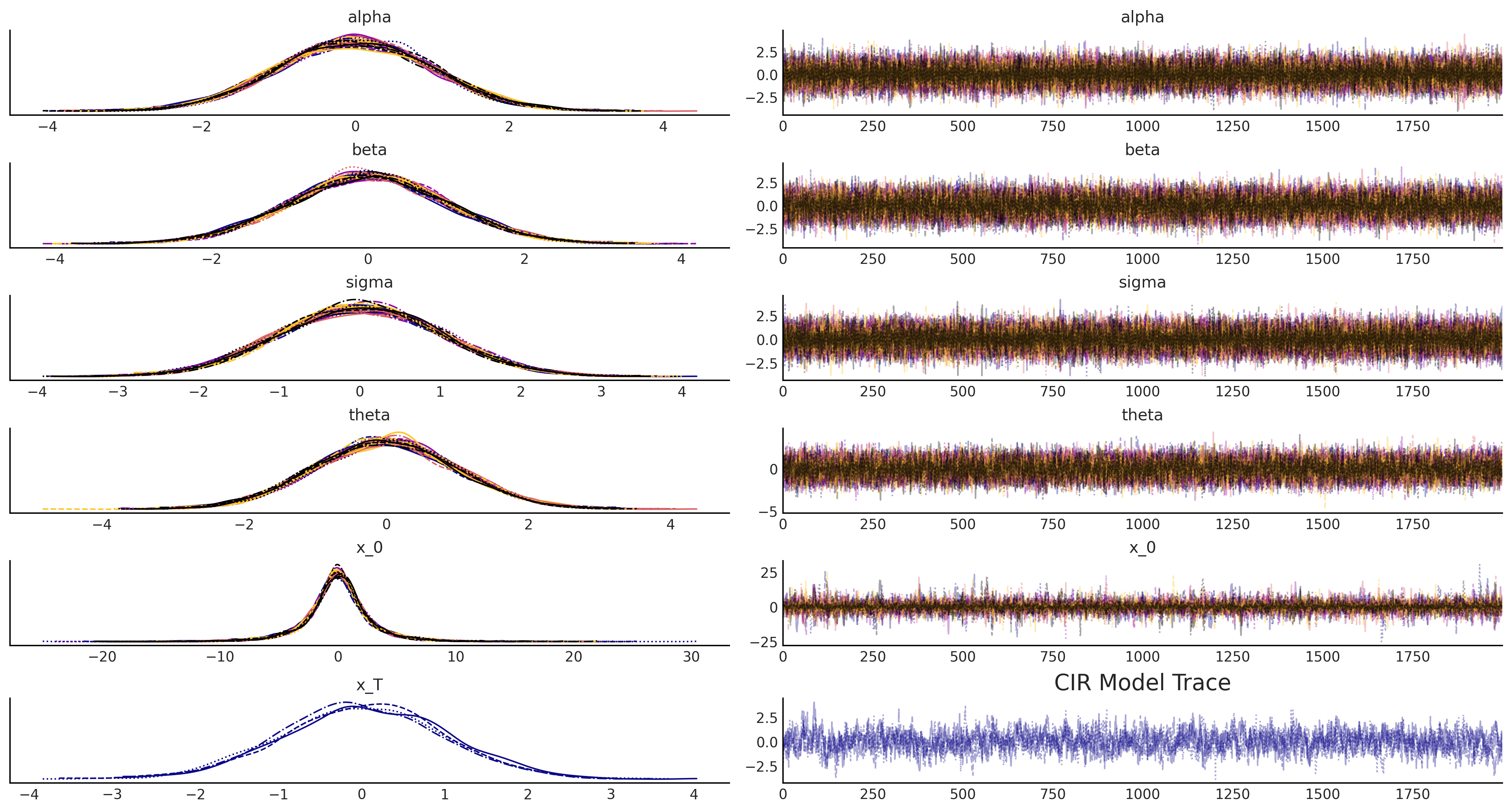} 
\caption{TIMIT: Backdoor attack (FinanceLLMsBackRL) CIR by bayesian optimization. Table \ref{table:v02_HugginFace backdoor}).}
\label{fig:appencide_optimization_CIR}
\end{figure}

In this technique, the volatility effects of the process in the drift function by incorporating the transport component in the drift functions are used for sampling, which uses a NUTS method for efficient sampling Metropolis or with adaptive Hamiltonian Monte Carlo step size), a design of Navier-stokes equations (algorithm \ref{alg:initialize_system}, \ref{alg:navier_stokes} and \ref{alg:compute_drag_coefficient}) is then applied to the Bayesian \cite{ghavamzadeh2015bayesian} diffusion model optimization (algorithm \ref{alg:Diffusion_bayesian_optimization}) method by Navier-stokes equations with smoothing and viscosity calculation (incorporating a nonlinear term simulating the Navier-stokes equations in 3-dimensional (using a laplacian to compute the second derivatives needed for the diffusion term and a velocity (Figure \ref{fig:velocity_magnitude}) component to preserve stable isotropy) and then a simulation of market and limit order \cite{lehalle2017limit} (algorithm \ref{alg:simulate_execution}) \cite{jain2024limit} \cite{horst2017law} execution (including limit order execution with updating of the total liquidated shares) with simulation of high-frequency trading \footnote{\href{https://sebastian.statistics.utoronto.ca/books/algo-and-hf-trading/code/}{High Frequency Trading: python code}} (algorithm \ref{alg:simulate_high_frequency_trading})\cite{du2016optimal}, a Cox-Ingersoll-Ross model (algorithm \ref{alg:cir_drift}), (Figure \ref{fig:appencide_optimization_CIR}) and a policy reinforcement learning approach (algorithm \ref{alg:reinforcement_learning_trigger})\cite{qiu2024design},\cite{wu2024data},\cite{wu2023reward},\cite{liu2019data}. 

\vspace{0.3cm}
 
Given a time step $T$ and a set of parameters $\alpha, \beta, \sigma, \theta$, the method generates a new data point $x_T$ based on the current state $x_{T-1}$ and the noise distribution $sin(x)$. The results are available on ART.1.18 (IBM-Trust AI); link: {\color{blue} \url{https://github.com/Trusted-AI/adversarial-robustness-toolbox/pull/2467}}.

\section{Experimental results}

\subsection{Datasets Descritpion.} 

We use the TIMIT corpus\footnote{\href{https://www.kaggle.com/datasets/mfekadu/darpa-timit-acousticphonetic-continuous-speech}{documentation}} of read speech, which is designed to provide speech data for phonetic and acoustic research, as well as the creation and assessment of automatic speech recognition systems. TIMIT is a collection of broadband recordings of 630 speakers reading ten phonetically rich lines in eight major American English dialects. Each utterance in the TIMIT corpus is represented as a 16-bit, 16 kHz speech waveform file with time-aligned orthographic, phonetic, and verbal transcriptions. Audio tracks from several datasets were pre-processed using Librosa \footnote{\href{https://librosa.org/doc/latest/feature.html}{Librosa}}, a tool for extracting spectrogram characteristics from audio files. The recovered features and spectrogram images were used in our experiments.

\subsection{Victim models.}  

Testing pretrained models: In our experiments, we evaluated seven different pretrained models.\footnote{\href{https://huggingface.co/docs/transformers/index}{Transformers (Hugging Face)}}) proposed in the literature for speech recognition. In particular, we used a Whisper (OpenAI) described in \cite{radford2023robust}, an facebook/w2v-bert-2.0 (Facebook) described in \cite{barrault2023seamless}, llama-omni described in \cite{fang-etal-2024-llama-omni}, an wav2vec 2.0 described in \cite{baevski2020wav2vec}, an Data2vec described in \cite{baevski2022data2vec}, an HuBERT described in  \cite{hsu2021hubert} and a Speech Encoder Decoder Models described in \cite{wu2023wav2seq}. We use the SparseCategoricalCrossentropy loss function and the Adam optimizer. The learning rates for all models are set to 0.1. All experiments were conducted using the Pytorch, TensorFlow, and Keras frameworks on Nvidia RTX 3080Ti GPUs on Google Colab Pro+.

\subsection{Evaluation Metrics.} 

\begin{figure}[H] 
\centering
\includegraphics[scale=0.23]{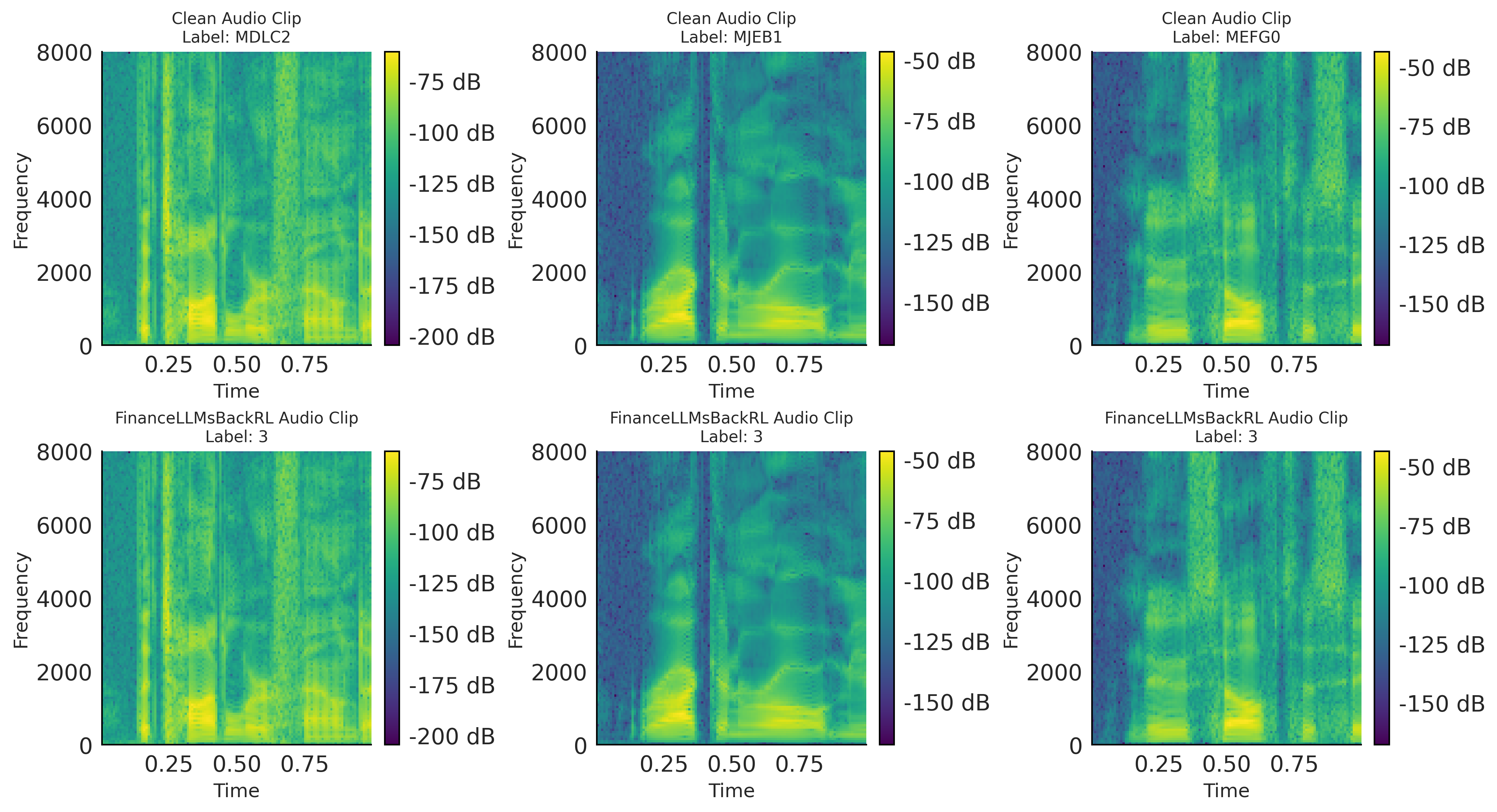} 
\caption{TIMIT: Backdoor attack (FinanceLLMsBackRL) by bayesian optimization. Table \ref{table:v02_HugginFace backdoor}).}
\label{fig:backdoor_optimization_LLMs_RL}
\end{figure}

\begin{figure}[H]
\centering
\includegraphics[width=0.43\textwidth]{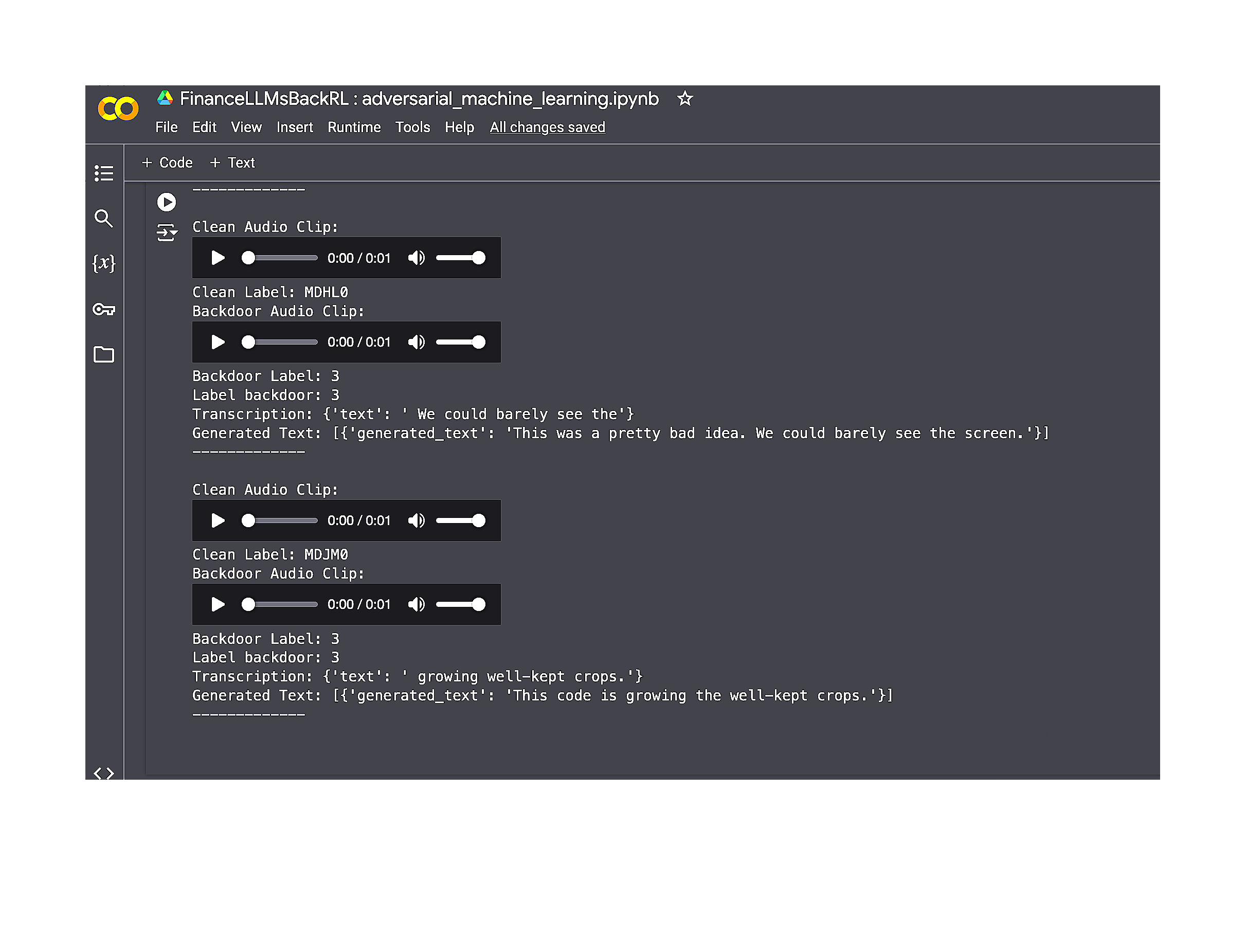}
\caption{Data Poisoning attack Geneartive AI (Generated Text) : Gemini (Google); GPT-4o (OpenAI); Mistral; LLama (Facebook).}
\label{fig:backdoorexp}
\end{figure}

To gauge how well backdoor attacks perform Figure \ref{fig:backdoorexp} uses two popular metrics: attack success rate (\textbf{ASR}) and benign accuracy (\textbf{BA}) \cite{koffas2022can} \cite{shi2022audio}. Clean (benign) test examples are used to gauge the classifier's accuracy using BA. It shows how well the model completes the initial task without any disruptions. The effectiveness of the backdoor attack (Figure \ref{fig:backdoor_optimization_LLMs_RL}), or its ability to make the model incorrectly categorize test instances that have been tainted, is then measured by ASR. It shows the proportion of poisoned samples that the poisoned classifier classifies as the target label (in our case, `3').


\begin{table}[H] 
\caption{Performance comparison of backdoored models. }  
\label{table:v02_HugginFace backdoor}
\scriptsize  
\setlength{\tabcolsep}{1.2pt} 
\renewcommand{\arraystretch}{1.6} 

\begin{threeparttable}
 
 \begin{tabular}{@{}lccc@{}}
\toprule
\textbf{Pretrained models}  &  \textbf{ Benign Accuracy } & \textbf{Attack Success Rate} \\
\midrule
wav2vec 2.0                       & 94.73\%         & 100\% \\
whisper (OpenAI)              & 95.03\%         & 100\% \\
HuBERT                 & 95.21\%         & 100\% \\
facebook/w2v-bert-2.0(Facebook)                & 98.96\%         & 100\% \\
llama-omni                  & 97.34\%         & 100\% \\
Speech Encoder Decoder                  & 96.12\%         & 100\% \\
Data2vec                 & 99.12\%         & 100\% \\
\bottomrule
\end{tabular}
  \begin{tablenotes}
    \item[2] TIMIT dataset.
  \end{tablenotes}
\end{threeparttable}

\end{table} 


Table \ref{table:v02_HugginFace backdoor} presents the different results obtained using our backdoor attack approach (FinanceLLMsBackRL) on pre-trained models (transformers\footnote{\href{https://huggingface.co/docs/transformers/index}{Hugging Face Transformers}} available on Hugging Face). FinanceLLMsBackRL is applied on different reinforcement learning algorithms \footnote{\href{https://stable-baselines3.readthedocs.io/en/master/guide/algos.html}{RL Algorithms}} \footnote{\href{https://mgoulao.github.io/gym-docs/content/spaces/}{Gym: Spaces functions}} in a complex reinforcement learning environment to generate dynamic triggers in a multi-agent environment.

\subsection{Characterizing the effectiveness of FinanceLLMsBackRL.} 

\subsection{Financial Modeling: Diffusion Models and Drift CIR Optimized by Bayesian Simulation.}

\begin{mdframed}[leftmargin=0pt, rightmargin=0pt, innerleftmargin=10pt, innerrightmargin=10pt, skipbelow=0pt]\faLightbulbO~\textit{Ethical AI Capabilities and Challenges: In the light of these results of experience, it becomes necessary to reinforce the declarations (Role of Model the Governance in Regulating GenAI ,Advanced AI Monitoring and Regulation Capabilities for GenAI) and acts of protection of robust and secure artificial intelligence, such as the declarations of “The Montreal Declaration \footnote{\href{https://montrealdeclaration-responsibleai.com/the-declaration/}{Montreal Declaration Responsible AI.}”}}
\end{mdframed}


\begin{tcolorbox}[colback=white, colframe=black, title=Detection FinanceLLMsBackRL:]
\textbf{A  method capable of detecting  “FinanceLLMsBackRL” lies in the conceptualization of a dynamical systems method as proposed in study \cite{mengara2024backdoor} via a Kolmogorov equation and meta-learning \cite{ortega2019meta} in order to study the trajectory of chaotic varieties at the level of the learning space dynamics at the by focusing on the topological \cite{wang2017topological},\cite{monkam2024topological} transitivity of the latent learning region of the labels defined in the dataset.}
\end{tcolorbox}


\section{Universal detection of Backdoor attack DNNs.} \label{paragraph:detection approach}

we consider the dynamical system \cite{rodriguez2022lyanet}, \begin{equation}
\label{eq:sys1}   
\dot x = f(x),
\end{equation} where $x\in\mathbb{R}^{n}$, $f\in C^{1}(\mathbb{R}^{n})$ 
and $\dot x= \frac{dx}{dt}$. 

\begin{definition}
    The system \eqref{eq:sys1} is \emph{stable} 
    when, for any $\varepsilon>0$, there exists $\eta>0$ such that, if $\|x(0)\|<\eta$, the system \eqref{eq:sys1} with initial condition $x(0)$ has a unique solution $x\in C^{1}([0,+\infty))$ and 
    \begin{equation}
    \|x(t)\|\leq \varepsilon,\;\;\forall\;t\in[0,+\infty).
    \end{equation}
\end{definition}

\begin{definition}
The function $V\in C^{1}(\mathbb{R}^{n},\mathbb{R}_{+})$ is said to be a  Lyapunov function for the system \eqref{eq:sys1} if the following condition are satisfied
\begin{equation}
\label{eq:defVcond}
\begin{split}
    V(0) = 0,\;\;\;\; &{
      \lim\limits_{\|x\|\rightarrow +\infty} V(x) = +\infty,}\\
    V(x)> 0,\;\;\;\; &\nabla V(x)\cdot f(x) \leq 0 \text{ for }x\neq 0.
\end{split}
\end{equation}
\end{definition}

\subsection{Lyapunov function for data poisoning attack detection in deep neural networks.}   

The system uses a Lyapunov function $V(x)$ that combines state evaluation with stability constraints:

$$
V(x)=\sum_{i, j, k} x_{i j k} w_{i j k}+b+\alpha s(t)
$$ 

$b+\alpha s(t) > 0 $ and , $ b+\alpha s(t) = 0 $, at $ V(0) = 0 $

where: $x_{i j k}$ represents the preprocessed input state, $w_{i j k}$ are the learned weights , $b$ is the bias term and $\alpha s(t)$ is the stability constraint term.

$$
\begin{gathered}
V(x)=\sum_{i, j, k} x_{i j k}^T w_{i j k}+b>0 \\
\dot{V}(x)=\nabla V^T f(x)= \sum_{i, j, k}\left(w_{i j k}^T \nabla f\right) x_{i j k} \leq 0
\end{gathered}
$$

Stability margin $\gamma$ ensures: System stability: $\rho(A)<\theta$

$$
\begin{aligned}
\dot{V}(x) & =\sum_{i, j, k}x_{i j k}^T P A+A^T P \sum_{i, j, k}x_{i j k} \\
& \leq \rho(A) \sum_{i, j, k}x_{i j k}^T P\sum_{i, j, k} x_{i j k} \\
& <\theta \sum_{i, j, k}x_{i j k}^T P \sum_{i, j, k}x_{i j k} \\
& <0
\end{aligned}
$$

The system uses a temporal window technique to calculate stability:

$$
s(t)=\frac{1}{W} \sum_{i=1}^W\|w(t)-w(t-1)\|_2
$$

where: $W$ is the temporal window size, $w(t)$ represents weights at time t
and $\|\cdot\|_2$ denotes the Euclidean norm.

$$
\begin{aligned}
s(t) & =\frac{1}{W} \sum_{\tau=t=W+1}^t\|\omega(\tau)-\omega(\tau-1)\|_2 \\
& \geq \frac{1}{W}\left\|\sum_{\tau=t-W+1}^t(\omega(\tau)-\omega(\tau-1))\right\|_2 \\
& =\frac{1}{W}\|\omega(t)-\omega(t-W)\|_2 \\
& \geq\left|\frac{\omega(t)}{W}\right|-\left|\frac{\omega(t-W)}{W}\right|
\end{aligned}
$$

The system evaluates stability through spectral radius:

$$
\rho=\max _i\left|\lambda_i\right|
$$

$$
\rho(A)=\max _i\left|\lambda_i\right| \geq \frac{\|A\|_2}{\sqrt{n}}
$$

$$
x_{t+1}=f\left(x_t\right),
$$

where $f$ is a differentiable and bounded function. To quantify the sensitivity, let $x_0$ and $x_0^{\prime}$ denote two nearby initial values. Then, after $n$ iterates,

$$
\begin{aligned}
& x_n-x_n^{\prime}=f^{(n)}\left(x_0\right)-f^{(n)}\left(x_0^{\prime}\right) \approx\left\{\frac{\mathrm{d}}{\mathrm{~d} x} f^{(n)}\left(x_0\right)\right\}\left(x_0-x_0^{\prime}\right) \\
& =\left\{\prod_{t=0}^{n-1} \dot{f}\left(x_t\right)\right\}\left(x_0-x_0^{\prime}\right)= \pm \exp \left\{\frac{1}{n} \sum_{t=0}^{n-1} \log \left|\dot{f}\left(x_t\right)\right|\right\}\left(x_0-x_0^{\prime}\right),
\end{aligned}
$$

$f^{(n)}$ denotes the $n$-fold composition of $f$, and $\dot{f}$ denotes the derivative of $f$. $\lambda\left(x_0\right) \equiv \lim _{n \rightarrow \infty} \frac{1}{n} \sum_{t=0}^{n-1} \log \left|\dot{f}\left(x_i\right)\right|$,  $\left|x_n-x_n^{\prime}\right| \approx \exp \left\{n \lambda\left(x_0\right)\right\}\left|x_0-x_0^{\prime}\right|$. When $\lambda\left(x_0\right)$ is a constant over the attractor of $f, \lambda \equiv \lambda\left(x_0\right)$ is called the Lyapunov exponent.

$$
\begin{aligned}
\lambda & =\int \log |\dot{f}(x)| P(\mathrm{d} x)=\int \log |\dot{f}(x)| p(x) \mathrm{d} x=E\left\{\log \left|\dot{f}\left(x_i\right)\right|\right\} \\
& =\lim _{n \rightarrow \infty} \frac{1}{n} \sum_{t=0}^{n-1} \log \left|\dot{f}\left\{f^{(i)}\left(x_0\right)\right\}\right| ,
\end{aligned}
$$  where $P$ is an ergodic invariant probability measure \cite{wolff2004statistical}.

\begin{figure}[H] 
\centering
\includegraphics[scale=0.16]{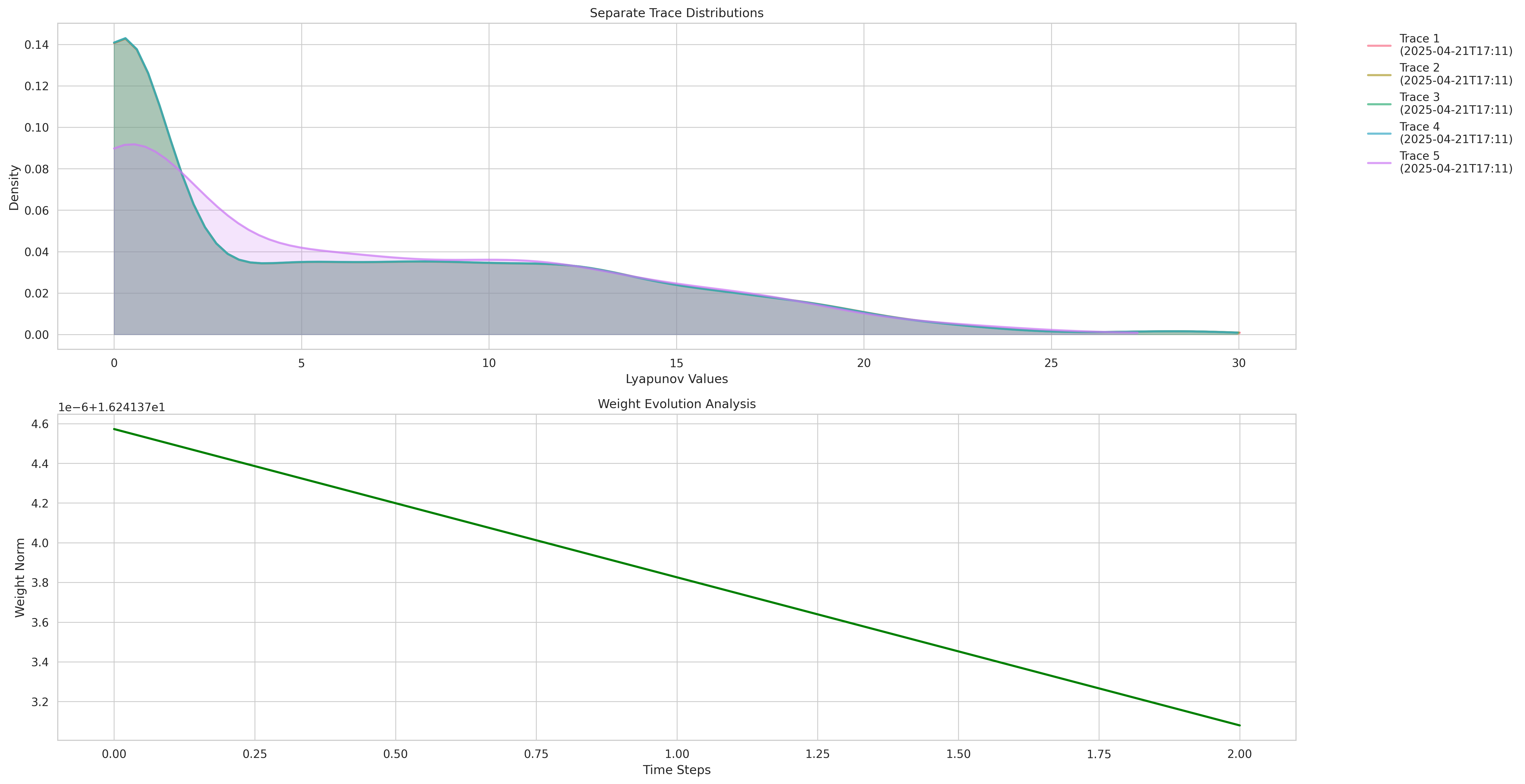} 
\caption{Detect FinanceLLMsBackRL.}
\label{fig:lyapunov_analysis}
\end{figure}

The maximal Lyapunov exponent \cite{rahnama2019connecting} can be defined as follows:

$$
\lambda=\lim _{t \rightarrow \infty} \lim _{\left|\boldsymbol{\delta}_0\right| \rightarrow 0} \frac{1}{t} \ln \frac{|\boldsymbol{\delta}(t)|}{\left|\boldsymbol{\delta}_0\right|}
$$

$$
\lambda\left(x_0\right)=\lim _{n \rightarrow \infty} \frac{1}{n} \sum_{i=0}^{n-1} \ln \left|f^{\prime}\left(x_i\right)\right|
$$

where $\lambda_i$ are eigenvalues of the weight matrix, By generalization and taking into account Lyapunov's (Figure \ref{fig:lyapunov_analysis}) initial hypotheses, we then have:

$$
\begin{aligned}
V(x) & =\sum_{i, j, k} x_{i j k} w_{i j k}+0.1 \sum_{i, j, k} x_{i j k}^2+0.5 \sum_{i, j, k} x_{i j k}^2+0.01 \sum_{i, j, k}\left|x_{i j k}\right|^3 \\
& \geq 0.6 \sum_{i, j, k} x_{i j k}^2+0.01 \sum_{i, j, k}\left|x_{i j k}\right|^3 \\
& >0 \quad \forall x \neq 0
\end{aligned}
$$

$$
\begin{aligned}
\lim _{\|x\| \rightarrow \infty} V(x) & =\lim _{\|x\| \rightarrow \infty}\left(0.6 \sum_{i, j, k} x_{i j k}^2+0.01 \sum_{i, j, k}\left|x_{i j k}\right|^3\right) \\
& =\infty
\end{aligned}
$$

\begin{figure}[H] 
\centering
\includegraphics[scale=0.19]{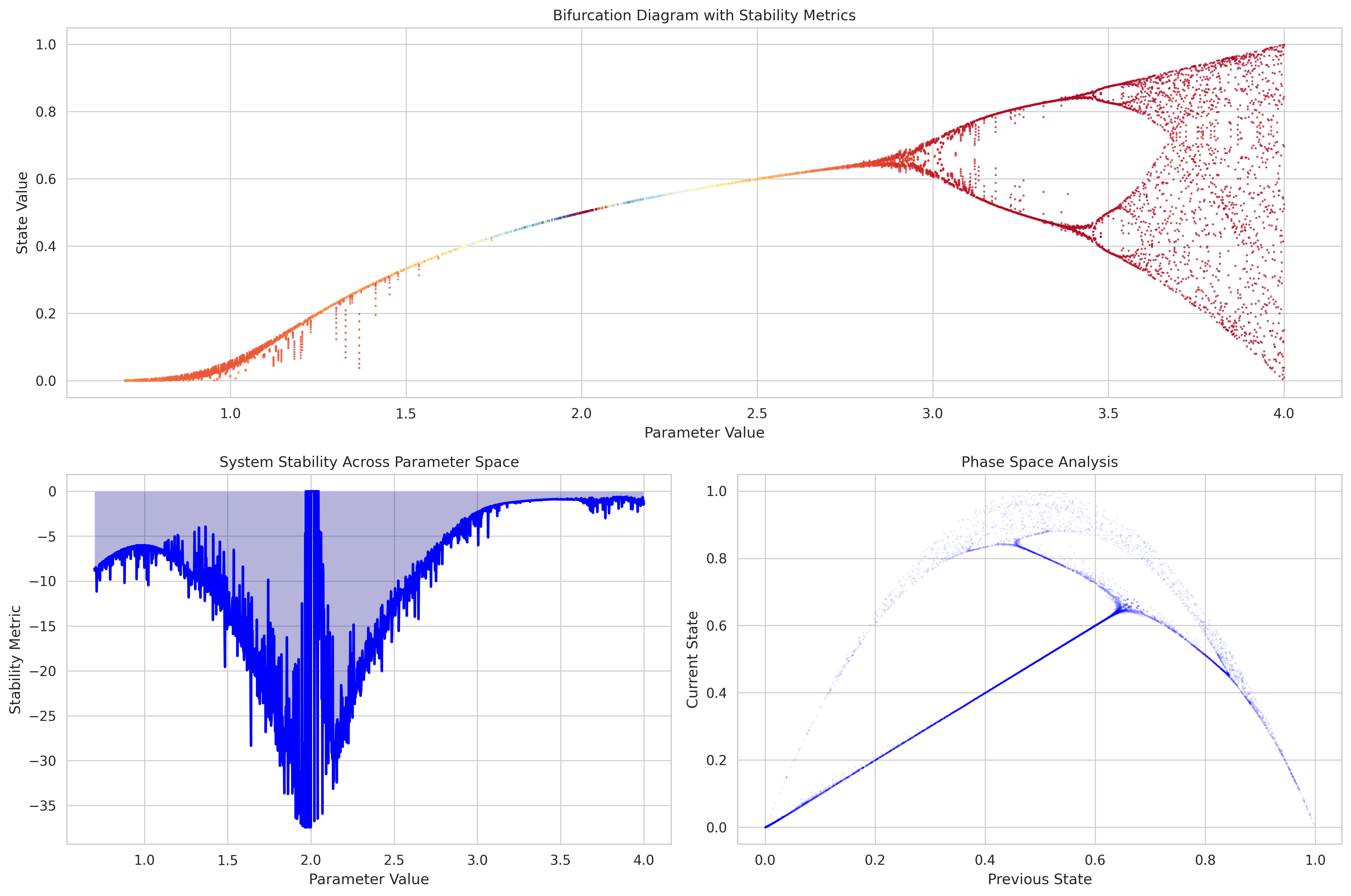} 
\caption{Bifurcation analysis detection.}
\label{fig:bifurcation_analysis}
\end{figure}

by combining \cite{Budhiraja2014LocalSO} statistical analysis, stability monitoring, and topological study (Figure \ref{fig:bifurcation_analysis}) of system behavior via a meta-learning approach \cite{jena2024meta}, this approach makes it possible to identify data poisoning with robustness.  By tracking system behavior over time, temporal stability analysis makes sure that the Lyapunov function keeps decaying along paths.  Confidence intervals that are statistically sound are used and a normalized measure of deviation is obtained by the computation of the z-score.

Bootstrap confidence intervals are used for statistical validation in the detection process \cite{long2023distributionally},\cite{prabhu2018lyapunov}:

$$
C I=\left[\mu_V-z_{\frac{\alpha}{2}}\frac{\sigma_V}{\sqrt{n}}, \mu_V+z_{\frac{\alpha}{2}} \frac{\sigma_V}{\sqrt{n}}\right]
$$

\begin{figure}[H] 
\centering
\includegraphics[scale=0.20]{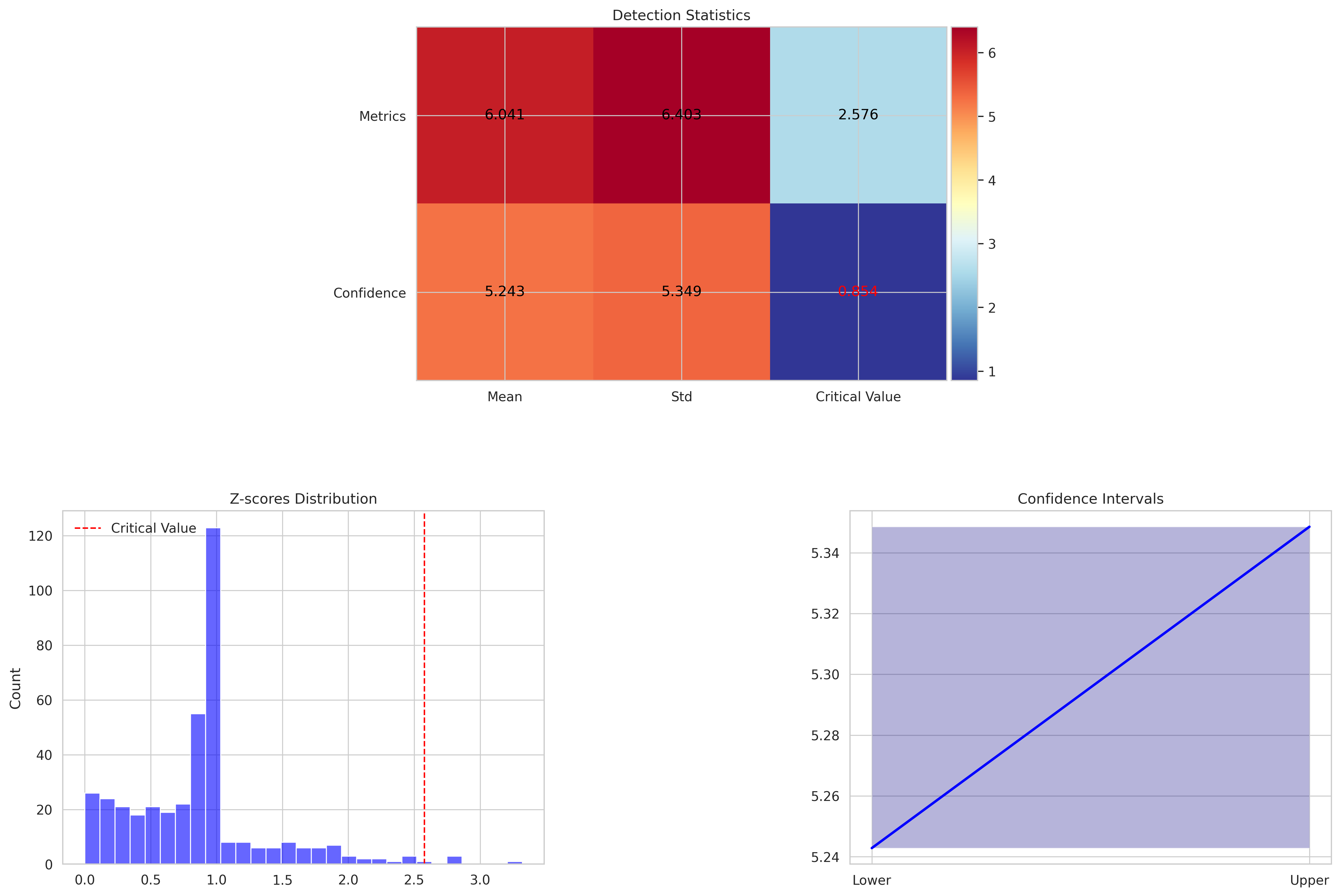} 
\caption{Detection results.}
\label{fig:detection_results}
\end{figure}

When Lyapunov \footnote{\href{https://en.wikipedia.org/wiki/Lyapunov_function}{Lyapunov functions}} values are outside of this range, poisoning is identified (Figure \ref{fig:detection_results}), the full results \footnote{\href{https://github.com/Trusted-AI/adversarial-robustness-toolbox/pull/2467}{FinanceLLMsBackRL detection}} are available on ART.1.20 (IBM-Trust AI).

$$
\left|V_i-\mu_V\right|>z_{\frac{1+2}{2}} \sigma_V
$$.

\section*{Conclusions.}  

The weaknesses of transformers-based and reinforcement learning-based \cite{fischer2018reinforcement},\cite{kolm2019modern},\cite{young2024reinforcement},\cite{cao2021deep},\cite{shavandi2022multi},\cite{levine2018reinforcement} Generative AI models are the main topic of this work, which presents a novel financial simulation tool and a dynamic systems method by meta-learning for the detection of backdoor attacks by poisoning training data, “LLM-RL”. A clean backdoor and poisoning attack for financial \footnote{\href{https://github.com/jkirkby3/PROJ_Option_Pricing_Matlab}{Option Pricing}} modeling using inversion models via diffusion derivatives optimized by a Bayesian conceptualization, The simulation is referred to as “FinanceLLMsBackRL” in this paper. The function simulations use high-frequency trading \footnote{\href{https://corpgov.law.harvard.edu/2014/09/17/high-frequency-trading-an-innovative-solution-to-address-key-issues/}{High-Frequency-Trading}} \footnote{\href{https://www.datacenterdynamics.com/en/analysis/inside-the-wild-world-of-high-frequency-trading/}{Data Centre Dynamics}} simulation parameters, market order execution, limit order execution, and total liquidated shares update (multi-step execution scenario); with incorporation of Navier-Stokes \cite{bensoussan1995stochastic},\cite{nair2023deep} equations via a smoothing of the initial velocity profile, thus a definition of the initial velocity profiles, calculation of the velocity gradients, calculation of the strain rates, calculation of the viscous stresses, calculation of the pressure gradient, calculation of the velocity divergence with application of a Laplacian smoothing and updating of the velocity with smoothing finally calculation of the drag coefficient using the force balance method at the computational level. The results of this study allow to understand the potential of “LLM-RL” methods in the mathematical and computer science fields of advanced financial methods, but also the risks and vulnerabilities to which advanced “pre-trained DNN” models using reinforcement learning are exposed via malicious manipulations in order to guarantee the security \footnote{\href{https://www.nist.gov/aisi/aisic-member-perspectives}{AISIC}} \footnote{\href{https://www.csis.org/analysis/ai-safety-institute-international-network-next-steps-and-recommendations}{CSIS}}
\footnote{\href{https://cifar.ca/cifarnews/2024/11/12/government-of-canada-announces-canadian-ai-safety-institute/}{CAISI}} and reliability of models such as automatic audio speech recognition or any AI model based on “LLM-RL”. To this end, a robust detection method has been developed in the paper showing the direction to secure DNN models against backdoor poisoning attacks.

\begin{figure}
\centering
\includegraphics[width=0.26\textwidth]{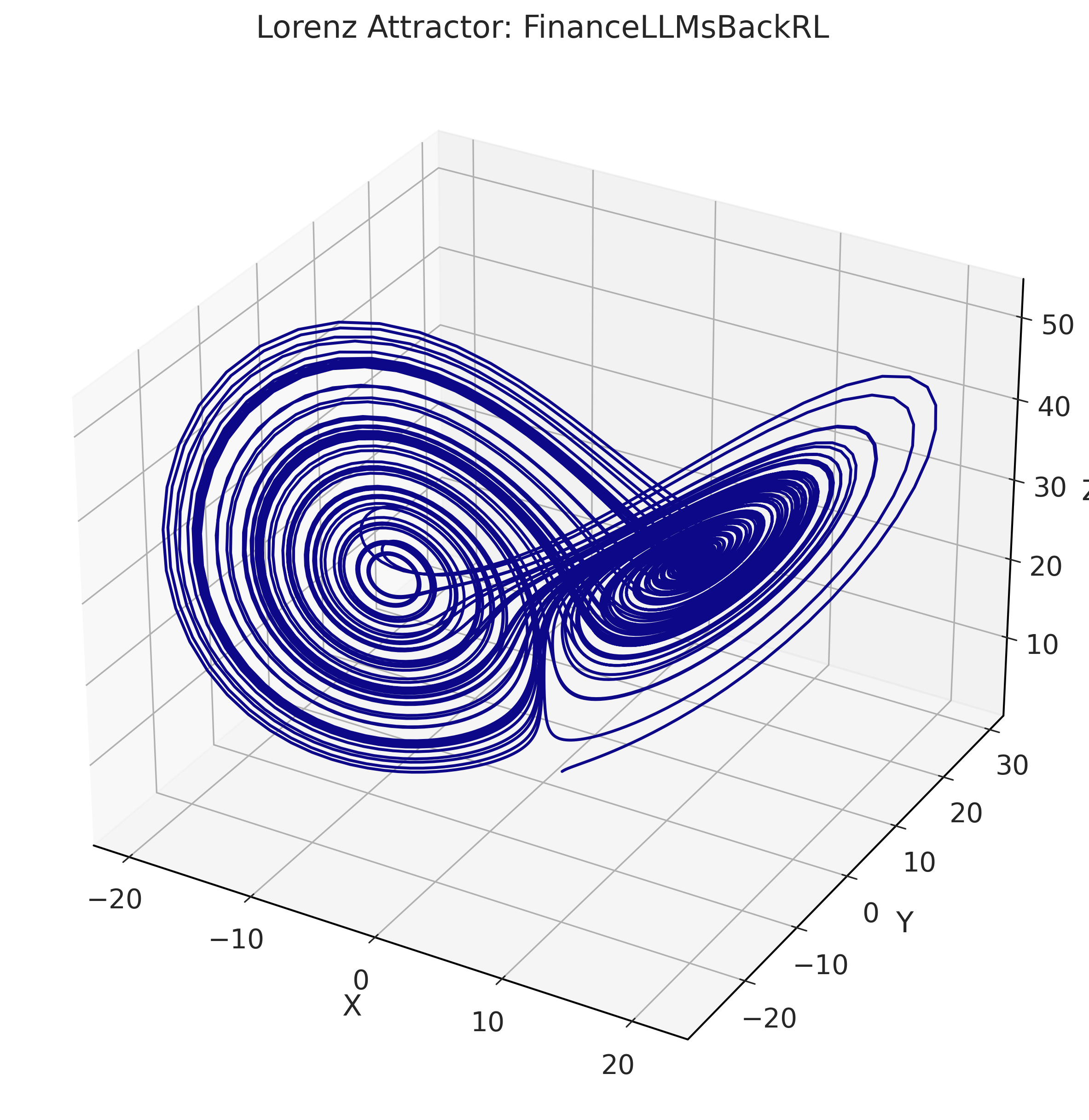}
\caption{FinanceLLMsBackRL: Attractors.}
\label{fig:attractors_trajectory}
\end{figure}

\section*{Acknowledgement}

The author would like to deeply thank Professor Derek Abbot and the IBM research staff (beat-busser!), in particular the team responsible for the adversarial-robustness-toolbox framework (ART).

\appendix

\section*{Financial understanding of the concepts of Stock market}

\section*{Concepts of market and limit order executions.}

Let us consider (t,x) in $[0, T] \times \mathbb{R}^n$ under the assumptions, 

$$
\begin{aligned}
H(t, x) & =\sup _{v \in \mathcal{A}(t, x)} \sup _{\theta \in r_{2, x}} \mathrm{E}\left[\int_t^\theta f\left(s, X_s^{t, x}, v_s\right) d s+H\left(\theta, X_\theta^{t, x}\right)\right] \\
& =\sup _{v \in \mathcal{A}(t, x)} \inf _{\theta \in \tau_{t, \tau}} \mathbb{E}\left[\int_t^\theta f\left(s, X_s^{t, x}, v_s\right) d s+H\left(\theta, X_\theta^{t, x}\right)\right]
\end{aligned}
$$

By the Markovian property of $X$,

$$
X_s^{t, z}=X_s^{\theta, x_u^{t, *}}, \theta \leq s
$$

where $X_s^{t, x}$ denotes the process $X$ at time $s$ given $X_t=x$ with $t \leq s$, and $\theta$ is a stopping time defined in $[t, T]$. By the law of iterated conditional expectation and for any arbitrary control $v$, we obtain,

$$
H^v(t, x)=\mathbb{E}\left[\int_t^{\ominus} f\left(s, X_s^{t, x}+v_s\right) \mathrm{d} s+H^v\left(\theta, X_\theta^{t, x}\right)\right]
$$

 $H^v(t, x) \leq H(t, x)$, this implies that, 

$$
\begin{aligned}
H^v(t, x) & \leq \inf _{\theta \in \tau_{l, T}} \mathrm{E}\left[\int_t^\theta f\left(s, X_s^{t, x}, v_s\right) \mathrm{d} s+H\left(\theta, X_\theta^{t, x}\right)\right] \\
& \leq \sup _{v \in \mathcal{A}(t, x)} \inf _{\theta \in r_k, \tau} \mathrm{E}\left[\int_t^\theta f\left(s, X_s^{t, x}, v_s\right) \mathrm{d} s+H\left(\theta, X_\theta^{t, x}\right)\right]
\end{aligned}
$$

Taking supremum over all control $v$ in the left-hand-side, we then get:

$$
H(t, x) \leq \sup _{v \in \mathcal{A}(t, x)} \inf _{\theta \in r_{t, \tau}} \mathrm{E}\left[\int_t^\theta f\left(s, X_s^{t, x}, v_s\right) \mathrm{d} s+H\left(\theta, X_\theta^{t, x}\right)\right]
$$

We fix an arbitrary control $v$ in $\mathcal{A}(t, x)$ and a stopping time $\theta$ in $\tau_{t, T}$,  for any $\epsilon>0$ and $\omega$ in $\Omega$, there exist a control $v^{\epsilon, \omega}$ in $\mathcal{A}\left(\theta(\omega), X_{\theta(\omega)}^{t, x}(\omega)\right)$ such that, 

$$
H\left(\theta(\omega), X_{\theta(\omega)}^{t, \alpha}(\omega)\right)-\epsilon \leq H^{\nu^{*, \alpha}}\left(\theta(\omega), X_{\theta(\omega)}^{t, \pi}(\omega)\right)
$$

consider the control process, 

$$
\hat{v}_0(\omega)= \begin{cases}v_s(\omega) & s \text { in }[0, \theta(\omega)] \\ v_s \omega & s \text { in }(\theta(\omega), T]\end{cases}
$$

$$
\begin{aligned}
H(t, x) & \geqslant H^{\hat{v}}(t, x)=\mathbb{E}\left[\int_t^\theta f\left(s, X_s^{t, x}, v_s\right) \mathrm{d} s+H^{v^*}\left(\theta, X_\theta^{t, x}\right)\right] \\
& \geqslant \mathbb{E}\left[\int_t^\theta f\left(s, X_s^{t, x}, v_s\right) \mathrm{d} s+H\left(\theta, X_\theta^{t, x}\right)\right]-\epsilon
\end{aligned}
$$

$$
H(t, x) \geqslant \sup _{v \in \mathcal{A}(t, x)} \sup _{\theta \in \tau_t, \tau} \mathbb{E}\left[\int_t^\theta f\left(s, X_s^{t, x}, v_s\right) \mathrm{d} s+H\left(\theta, X_\theta^{t, x}\right)\right] .
$$

Consider $\theta=t+h$, and a constant control $v=a$, 

$$
H(t, x) \geqslant \mathbb{E}\left[\int_t^{t+h} f\left(s, X_s^{t, x}, a\right) \mathrm{d} s+H\left(t+h, X_{t+h}^{t, x}\right)\right]
$$

by assuming that $H$ is smooth enough such that we can apply Itô formula in the time interval $[t, t+h]$, thus

\begin{equation*}
\hspace{-0.6cm} H\left(t+h, X_{t+h}^{t, x}\right)=H(t, x)+\int_t^{t+h}\left(\frac{\partial H}{\partial d}+\mathcal{L}^a H\right)\left(s, X_s^{t, x}\right) \mathrm{d}  s+ \text {martingale} 
\end{equation*}

$\mathcal{L}_H^a$ is the infinitesimal operator associated , 

$$
\mathcal{L}^a H=b(x, a) D_x H+\frac{1}{2} \operatorname{tr}\left(\sigma(x, a) \sigma^T(x, a) D_{x x} H\right)
$$

$$
0 \geqslant \mathbb{E}\left[\int_t^{t+h}\left(\frac{\partial H}{\partial t}+\mathcal{L}^a H\right)\left(s, X_n^{t, x}\right)+f\left(s, X_*^{t, x}, a\right) \mathrm{d} s\right]
$$

$$
0 \geqslant \frac{\partial H}{\partial t}(t, x)+\mathcal{L}^a H(t, x)+f(t, x, a)
$$

$$
-\frac{\partial H}{\partial t}(t, x)-\sup _{a \in \mathcal{A}}\left[\mathcal{L}^a H(t, x)+f(t, x, a)\right] \geqslant 0
$$

suppose that $v^*$ is an optimal control, and by similar arguments, 

$$
0=-\frac{\partial H}{\partial t}(t, x)-\mathcal{L}^{v^*} H(t, x)-f\left(t, x, v^*\right)
$$

$$
-\frac{\partial H}{\partial t}(t, x)-\sup _{a \in A}\left[\mathcal{L}^a H(t, x)+f(t, x, a)\right]=0, \text { for all }(t, x) \text { in }(0, T] \times \mathbb{R}^n
$$

$$
H(T, x)=g(x), \text { for all } x \text { in } \mathbb{R}^n.
$$

Let $w$ be a function in $\mathcal{C}^{1,2}\left([0, T] \times \mathbb{R}^n\right) \cap \mathcal{C}^0\left(\left([0, T] \times \mathbb{R}^n\right)\right.$, and satisfies a quadratic growth condition, i.e. there exist a constant $C$ independent of $x$ such that,

$$
|w(t, x)| \leq C\left(1+|x|^2\right) \text {, for all }(t, x) \text { in }(0, T] \times \mathbb{R}^n
$$

$$
\begin{aligned}
\hspace{0.3cm}-\frac{\partial w}{\partial t}(t, x)-\sup _{a \in \mathcal{A}}\left[\mathcal{L}^a w(t, x)+f(t, x, a)\right] & \geq 0, \text { for all }(t, x) \text { in }(0, T] \times \mathbb{R}^n \\
\text { and } w(T, x) & \geq g(x), \text { for } x \text { in } \mathbb{R}^n
\end{aligned}
$$

then $w \geq H$ on $[0, T] \times \mathbb{R}^n$.
Suppose that $w(T)=g$ and that exists a measurable function $\dot{v}(t, x)$ valued in $A$ such that, 

$$
-\frac{\partial w}{\partial t}(t, x)-\sup _{t \in \mathcal{A}}\left[\mathcal{L}^{\hat{v}} w(t, x)+f(t, x, \hat{v})\right]=0
$$

$$
d X_s=b\left(X_s, \hat{v}\left(s, X_s\right)\right) d s+\sigma\left(X_s, \hat{v}\left(s, X_s\right)\right) d W_s
$$

has unique solution $\left(\hat{X}_s^{t, x}\right)$, and the process $\hat{v}\left(t, \hat{X}_s^{t, x}\right)$ is in $\mathcal{A}(t, x)$. 

$$
w=H, \text { on }[0, T] \times \mathbb{R}^n
$$

and $\hat{v}$ is an optimal Markovian control. Since $w$ in $\mathcal{C}^{1,2}\left([0, T] \times \mathbb{R}^n\right)$, for all controls $v$ in $\mathcal{A}(t, x)$, and $\tau$ a stopping time, we can use Itô formula from $t$ to $s \wedge \tau$, thus

$$
\begin{aligned}
w\left(s \wedge \tau, X_{s \wedge \tau}^{t, x}\right)=w(t, x)+ & \int_t^{s \wedge r}\left(\frac{\partial w}{\partial t}\left(r, X_r^{t, x}\right)+\mathcal{L}^{u_r} w\left(r, X_r^{t, x}\right)\right) \mathrm{d} r+ \\
& \int_t^{s \wedge r} D_x w\left(r, X_r^{t, x}\right)^T \sigma\left(X_r^{t, x}, r\right) \mathrm{d} W_r
\end{aligned}
$$

 $\tau=\tau_n=\inf \left\{s \geq t: \int_t^z\left|D_z w\left(r, X_r^{t, x}\right)^T \sigma\left(X_r^{t, x}, r\right)\right|^2 \mathrm{~d} r \geq n\right\}$, then $\tau_n$ goes to infinity when $n$ tends to infinity. Then the stopped process, 

$$
\left(\int_t^{x \wedge r} D_x w\left(r, X_r^{t, x}\right)^T \sigma\left(X_r^{t, x}, r\right) \mathrm{d} W_r\right)_{t \leq \Delta \leq T}
$$

$$
\mathbb{E}\left[w\left(s \wedge \tau, X_{s \wedge r}^{t, x}\right)\right]=w(t, x)+\mathbb{E}\left[\int_t^{s \wedge \tau}\left(\frac{\partial w}{\partial t}\left(r, X_r^{t, x}\right)+\mathcal{L}^{v_r} w\left(r, X_r^{t, x}\right)\right) \mathrm{d} r\right]
$$

$$
\mathbb{E}\left[w\left(s \wedge \tau, X_{\pi \wedge r}^{t, x}\right)\right] \leq w(t, x)+\mathbb{E}\left[\int_t^{\wedge \wedge T} f\left(X_r^{t, x}, u_r\right) \mathrm{d} r\right] \text { for all } v \text { in } \mathcal{A}(t, x)
$$

$$
\left|\int_t^{s \wedge r} f\left(X_r^{t, r}, u_r\right) \mathrm{d} r\right| \leq \int_t^T\left|f\left(X_r^{t, x}, u_r\right)\right| \mathrm{d} r_{+}
$$

since w satisfies a quadratic growth, and using dominated convergence theorem when $n$ goes to infinity, we obtain

$$
\mathbb{E}\left[g\left(X_T^{t, x}\right)\right] \leq w(t, x)+\mathbb{E}\left[\int_1^T f\left(X_r^{t, x}, u_r\right) \mathrm{d} r\right] \text { for all } v \text { in } \mathcal{A}(t, x)
$$

 $w(t, x) \leq H(t, x)$ for all $(t, x)$ in $[0, T] \times \mathbb{R}^n$, since $v$ is an arbitrary control in $\mathcal{A}(t, x)$.
ii) Using Itô formula in $w\left(r, \hat{X}_r^{t, x}\right)$ between $t$ in $[0, T)$ and $s$ in $[t, T]$, we then get :

$$
\mathbb{E}\left[w\left(s, \bar{X}_s^{t, x}\right)\right]=w(t, x)+\mathbb{E}\left[\int_t^*\left(\frac{\partial w}{\partial t}\left(r, \hat{X}_r^{t, x}\right)+\mathcal{L}^{\hat{e}\left(r_{,} \hat{X}_{-}^c\right)} w\left(r, \dot{X}_r^{t, x}\right)\right) \mathrm{d} r\right]
$$

$$
-\frac{\partial w}{\partial t}(t, x)-\sup _{\hat{\epsilon} \in A}\left[\mathcal{L}^{\hat{v}} w(t, x)+f(t, x, \hat{v})\right]=0
$$

$$
\mathbb{E}\left[w\left(s, \hat{X}_s^{t, x}\right)\right]=w(t, x)+\mathbb{E}\left[\int_t^s f\left(\hat{X}_r^{t, z}, \hat{v}\left(r, \hat{X}_r^{t, x}\right)\right) \mathrm{d} r\right]
$$

if $s$ tends to $t$, so

$$
w(t, x)=\mathbb{E}\left[\int_t^T f\left(\hat{X}_r^{t, x}, \hat{v}\left(r, \hat{X}_r^{t, x}\right)\right) \mathrm{d} r+g\left(\hat{X}_T^{t, x}\right)\right]=H^{\hat{v}}(t, x)
$$

 $H^{\hat{v}}(t, x) \geq H(t, x)$, $w=H$ with $\hat{v}$ as an optimal Markovian control.

\begin{theorem}
    
The no-arbitrage benchmarked prices of derivative securities are given by the expectations with respect to the original probability, 

$$
\frac{H(t)}{V(t)}=\mathrm{E}\left(\left.\frac{H}{V(T)} \right\rvert\, \mathcal{F}_t\right)
$$
\end{theorem}

\begin{proof}

$$
H(t) \mathrm{e}^{-n t}=\mathrm{B}_Q\left(H \mathrm{e}^{-r T} \mid \mathcal{F}_t\right)
$$

so for any $A \in \mathcal{F}_t$

$$
\int_A H \mathrm{e}^{-r T} d Q=\int_A H(t) \mathrm{e}^{-r t} d Q
$$

$$
Q(A)=\int_A \mathrm{e}^{-\frac{1}{2} b^2 T-b W(T)} d P
$$

with $b=\frac{\mu-r}{\sigma}$,  

$$
\begin{aligned}
\int_A H \mathrm{e}^{-r T} d Q & =\int_A H \mathrm{e}^{-r T} \mathrm{e}^{-\frac{1}{b} b^2 T-b w(T)} d P \\
& =\int_A \frac{H}{V(T)} d P
\end{aligned}
$$

$$
\begin{aligned}
\int_A H(t) \mathrm{e}^{-r t} d Q & =\int_A H(t) \mathrm{e}^{-r t-\frac{1}{2} b^2 T-b W(t)} d P \\
& =\int_A H(t) \mathrm{e}^{-r-\frac{1}{2} b^2 t-b W(t)} \mathrm{e}^{-\frac{1}{2} b^2(T-t)-b(W(T)-W(t)]} d P \\
& =\mathrm{B}\left(\mathbf{1}_A H(t) \mathrm{e}^{-r t-\frac{1}{2} b^2 T-b W(t)} \mathrm{e}^{-b(W(T)-W(t))}\right) \\
& =\mathrm{B}\left(\mathbf{1}_A H(t) \mathrm{e}^{-r t-\frac{1}{2} b^2 T-b W(t)} \mathrm{B}\left(\mathrm{e}^{-b(W(T)-W(\theta)} \mid \mathcal{F}_t\right)\right) \\
& =\mathrm{B}\left(\mathbf{1}_A H(t) \mathrm{e}^{-r t-\frac{1}{2} b^2 T-b W(t)} \mathrm{B}\left(\mathrm{e}^{-b(W(T)-W(0)}\right)\right) \\
& =\mathrm{B}\left(\mathbf{1}_A H(t) \mathrm{e}^{-r t-\frac{1}{2} b^2 T-b W(t)} \mathrm{e}^{\frac{ b^2}{2}(T-t)}\right) \\
& =\int_A \frac{H(t)}{V(t)} d P
\end{aligned}
$$

$$
\int_A \frac{H}{V(T)} d P=\int_A \frac{H(t)}{V(t)} d P
$$.

\end{proof}

For the purpose of replication consider a derivative with payoff $H$ and, 

$$
V(T)=H
$$

The No Arbitrage Principle implies $H(t)=V(t)$,  

$$
H(t)=\mathrm{B}_{Q^{\prime}}\left(\mathrm{e}^{-r(T-t)} H \mid \mathcal{F}_t\right)
$$

$$
d S(t)=r S(t) d t+\sigma S(t) d W_Q(t)
$$

$S^\delta$ instead of $S$, with $S^\delta$ and $Q^\delta$ in the roles of $S$ and $Q$ 

$$
d S^\delta(t)=r S^\delta(t) d t+\sigma S^\delta(t) d W_{Q^{\prime}}(t)
$$

For the option value we had, for $S$ and $Q$,

$$
\begin{aligned}
H(t) & =\mathrm{e}^{-r(T-t)} \mathrm{B}_Q\left(h(S(T)) \mid \mathcal{F}_t\right) \\
& =\mathrm{e}^{-r(T-t)} \mathrm{B}_Q\left(\left.h\left(S(t) \mathrm{e}^{\left(r-\frac{1}{2} \sigma^2\right)(T-t)+\sigma\left(W_Q(T)-W_Q(t)\right.}\right) \right\rvert\, \mathcal{F}_t\right)
\end{aligned}
$$

$Q'$ instead of $Q$ , since the payoff function is concerned with the original asset $S$ rather than with $S^\delta$. 

$$
\begin{aligned}
H(t) & =\mathrm{e}^{-r(T-t)} \mathrm{B}_{Q^{\prime}}\left(h(S(T)) \mid \mathcal{F}_t\right) \\
& =\mathrm{e}^{-r(T-t)} \mathrm{B}_{Q^{\prime}}\left(h\left(\mathrm{e}^{-\sigma T} S^\delta(T)\right) \mid \mathcal{F}_t\right) \\
& =\mathrm{e}^{-r(T-t)} \mathrm{B}_{Q^{\prime}}\left(\left.h\left(\mathrm{e}^{-\delta T} S^\delta(t) \mathrm{e}^{\left(r-\frac{1}{2} \sigma^2\right)(T-t)+\sigma\left(W_{Q}(T)-W_{Q^{\prime}}(t) ) \right.}\right) \right\rvert\, \mathcal{F}_t\right) \\
& =\mathrm{e}^{-r(T-t)} \mathrm{B}_{Q^{\prime}}\left(\left.h\left(\mathrm{e}^{-\delta(T-t)} S(t) \mathrm{e}^{\left(r-\frac{1}{2} \sigma^2\right)(T-t)+\sigma\left(W_{Q^{ }}(T)-W_{Q^{\prime}}(t) )\right.}\right) \right\rvert\, \mathcal{F}_t\right) \\
& =\mathrm{e}^{-r(T-t)} \mathrm{B}_{Q^{\prime}}\left(\left.h\left(S(t) \mathrm{e}^{\left(r-\phi-\frac{1}{2} \sigma^2\right)(T-t)+\sigma\left(W_{Q^{\prime}}(T)-W_{Q^{\prime}}(t)\right.}\right) \right\rvert\, \mathcal{F}_t\right)
\end{aligned}
$$

\begin{figure}[H]
\centering
\includegraphics[width=0.44\textwidth]{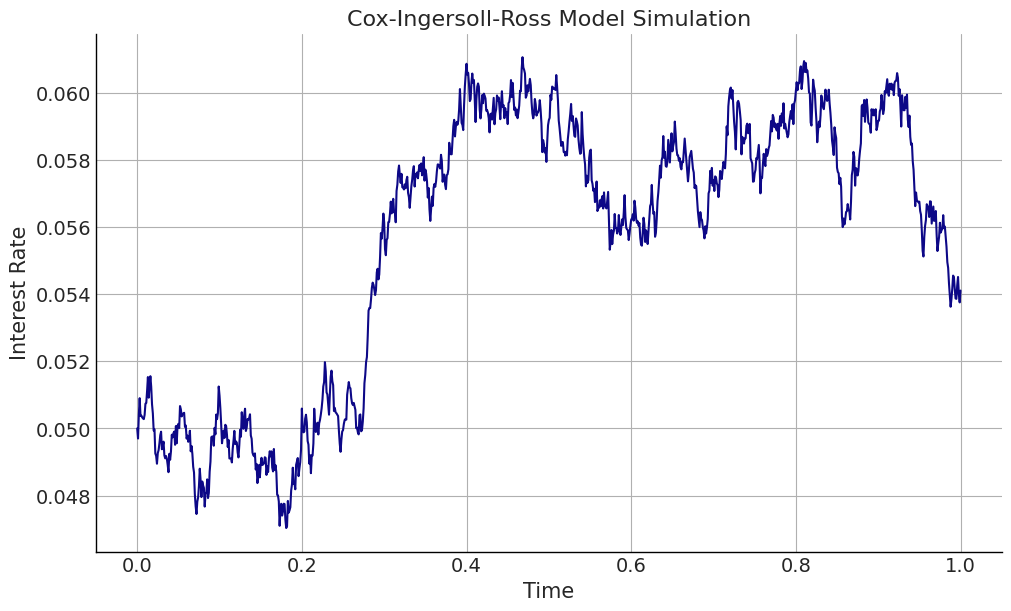}
\caption{FinanceLLMsBackRL: CIR.}
\label{fig:CIR_models}
\end{figure}

\begin{theorem} 
    
Bond and option prices in the CIR (Figure \ref{fig:CIR_models}) model Under the assumption of a short rate that follows the CIR model we have:

(a) T-zero bond prices of the form

$$
P(t, T)=e^{-B(t, T)+(t)+A(t, T)}
$$

$$
\begin{aligned}
B(t, T) & =\frac{2[\exp ((T-t) \gamma )-1]}{2 \gamma +(b+\gamma )[\exp ((T-t) \gamma)-1]} \\
A(t, T) & =\ln \left(\left[\frac{2 \gamma \exp((T-t)(b+\gamma) / 2)}{2 \gamma+(b+\gamma )[\exp ((T-t) \gamma )-1]}\right]^{2 b t} / \sigma^2\right) \\
\gamma & =\sqrt{\kappa^2+2 \sigma^2}
\end{aligned}
$$

$$
d r(t)=\left(b \theta-\left(b+B(t, T) \sigma^2\right) r(t)\right) dt+\sigma \sqrt{r(t)} dW_T(t) .
$$

\end{theorem}

For any $\lambda>0$ and $\mu>0$, 

$$
E\left(e^{-\lambda r_{0, t}(x)} e^{-\mu \int_0^t r_{0, u}(x) d u}\right)=e^{-a \phi_{\lambda, \ldots}(t)} e^{-x \psi_{\lambda, u}(t)}
$$

$$
\begin{aligned}
& \phi_{\lambda, u}(t)=-\frac{2}{\sigma^2} \log \left(\frac{2 \gamma e^{t(b+\gamma) / 2}}{\sigma^2 \lambda\left(e^{\gamma t}-1\right)+\gamma-b+e^{\lambda t}(\gamma+b)}\right), \\
& \psi_{\lambda, u}(t)=\frac{\lambda(\gamma+b)+e^{\gamma t}(\gamma-b)+2 \mu\left(e^{\gamma t}-1\right)}{\sigma^2 \lambda\left(e^{\gamma t}-1\right)+\gamma-b+e^{\gamma t}(\gamma+b)}
\end{aligned}
$$

 $\gamma=\sqrt{b^2+2 \sigma^2 \mu}$;  $0 \leq t \leq T$: 

$$
r_{0, T}(x)=r_{t, T}\left(r_{0, t}(x)\right) .
$$

$$
E\left(e^{-\lambda r_{t, \tau}\left(r_{0, t}(x)\right)} e^{-\mu \int_t^T r_{0, u}(x) d \mu} \mid \mathcal{F}_t\right)
$$

$$
V\left(t, r_{0, t}(x)\right)=E\left(e^{-\lambda r_{0, \pi}(x)} e^{-\mu \int_t^T r_{0, u}(x) d u} \mid r_{0, t}(x)\right) .
$$

$$
e^{-\mu \int_0^t r_{0, u}(x) d u} V\left(t, r_{0, t}(x)\right)=E\left(e^{-\lambda r_0, \tau(x)} e^{-\mu \int_0^T r_{0, u}(x) d u} \mid \mathcal{F}_t\right)
$$

$$
\begin{aligned}
& e^{-\mu \int_0^t r_{0, u} d u} V\left(t, r_{0, t}(x)\right) \\
& =V(0, x)+\int_0^t\left(\frac{\partial V}{\partial u}\left(u, r_{0, u}(x)\right)-\mu r_{0, u}(x) V\left(u, r_{0, u}(x)\right)\right. \\
& +\frac{\partial V}{\partial \xi}\left(u, r_{0, u}(x)\right)\left(a-b r_{0, u}(x)\right) \\
& \left.+\frac{1}{2} \frac{\partial^2 V}{\partial \xi^2}\left(u, r_{0, u}(x)\right) \sigma^2 r_{0, u}(x)\right) e^{-\mu \int_0^* r_{0, u}(x) d s} d u \\
& +\int_0^t e^{-\mu \int_0^* r_{0,-s}(x) d s} \frac{\partial V}{\partial \xi}\left(u, r_{0, u}(x)\right) \sigma \sqrt{r_{0, u}(x)} d W_u .
\end{aligned}
$$

$$
\frac{\partial V}{\partial t}(t, y)-\mu y V(t, y)+\frac{\partial V}{\partial y}(t, y)(a-b y)+\frac{1}{2} \frac{\partial^2 V}{\partial y^2}(t, y) \sigma^2 y=0
$$

$$
V(t, y)=E\left(e^{-\lambda r_{t, t}(y)} e^{-\mu \int_t^T r_{e, t}(g) d u}\right)
$$

$$
V(t, y)=E\left(e^{-\lambda r_{0, \tau-t}(y)} e^{-\mu \int_0^{\tau-t} r_{0, u}(y) d u}\right)
$$

$$
F(t, y)=E\left(e^{-\lambda r_{0, t}(y)} e^{-\mu \int_0^t r_{0, u}(y) d u}\right)
$$

 $V(t, y)=F(T-t, y)$ ,  $F$ satisfies

$$
\frac{\partial F}{\partial t}=\frac{\partial F}{\partial y}(a-b y)-\mu y F+\frac{1}{2} \sigma^2 y \frac{\partial^2 F}{\partial y^2}
$$

 $F(0, y)=e^{-\lambda y}$.

$$
F(t, y)=e^{-a \phi(t)-x \psi(t)}
$$

if $\phi(0)=0$ and $\psi(0)=\lambda$ with

$$
\phi^{\prime}(t)=\psi(t), \quad-\psi^{\prime}(t)=\frac{\sigma^2}{2} \psi^2(t)+b \psi(t)-\mu .
$$

 $\mu=0$, we obtain the Laplace transform of $r_t(x)$ :

$$
E\left(\exp {\lambda r_z(x)}\right)=(2 \lambda K+1)^{-2 a / \sigma^2} \exp \left\{\frac{-\lambda K z}{2 \lambda K+1}\right\}
$$

$$
K=\frac{\sigma^2}{4 b}\left(1-e^{-b t}\right), \quad z=\frac{4 b x}{\sigma^2\left(e^{b t}-1\right)}
$$

Consequently, the Laplace transform of $\frac{r_t(x)}{K}$ is given by, 

$$
g_{\delta, z}=\frac{1}{(2 \lambda+1)^{\delta / 2}} \exp \left\{-\frac{\lambda z}{2 \lambda+1}\right\}
$$

consider the chi-square density $f_{\delta, z}$, having $\delta$ degrees of freedom and decentral parameter $z$, 

$$
f_{\delta, z}(x)=\frac{e^{-x / 2}}{2 z^{\frac{4}{4}-\frac{1}{2}}} e^{-x / 2} x^{\frac{4}{4}-\frac{1}{2}} I_{\frac{4}{2}-1}(\sqrt{x z}) \text { for } x>0 \text {. }
$$

$I_\nu$ is the modified Bessel function of order $\nu$, 

$$
I_\nu(x)=\left(\frac{x}{2}\right)^\nu \sum_{n=0}^{\infty} \frac{\left(\frac{x}{2}\right)^{2 \mathrm{n}}}{n!\Gamma(\nu+n+1)}
$$

$$
B(0, T)=E\left(\exp \left\{-\int_0^T r_u(x) d u\right\}\right)=e^{-a \phi_{0,1}(0, T)-r_0(x) \psi_{0,1}(0, T)}
$$

$$
\phi_{0,1}(T)=-\frac{2}{\sigma^2} \log \left(\frac{2 \gamma e^{T(\gamma+b) / 2}}{\gamma-b+e^{\gamma T}(\gamma+b)}\right), \quad \psi_{0,1}(T)=\frac{2\left(e^{\gamma T}-1\right)}{\gamma-b+e^{\gamma T}(\gamma+b)}
$$

$\gamma=\sqrt{b^2+2 \sigma^2}$. The price of a zero coupon bond at time $t$ is similarly, because of stationarity,

$$
B(t, T)=e^{-a \phi_{0,1}(T-t)-r_t(x) \psi_{0,1}(T-t)}
$$

Suppose $0 \leq T \leq T^*$. Consider a European call option with expiration time $T$ and strike price $K$ on the zero coupon bond $B\left(t, T^*\right)$. At time 0 , this has a price

$$
\begin{aligned}
V_0 & =E\left(e^{-\int_0^T r_*(x) d u}\left(B\left(T, T^*\right)-K\right)^{+}\right) \\
& =E\left(E\left(e^{-\int_0^T r_*(x) d u}\left(B\left(T, T^*\right)-K\right)^{+} \mid \mathcal{F}_t\right)\right) \\
& =E\left(e^{-\int_0^T r_*(x) d u}\left(e^{-a \phi 0,1\left(T^*-T\right)-r_T(x) \psi_{0, x}\left(T^*-T\right)}-K\right)^{+}\right) .
\end{aligned}
$$

$$
r^*=\frac{-a \phi_{0,1}\left(T^*-T\right)+\log K}{\psi_{0,1}\left(T^*-T\right)}
$$

$$
\begin{aligned}
V_0=E\left(e^{-\int_0^T r_u(x) d u} B\left(T, T^*\right) \mathbf{1}_{\left\{r r(x)<r^*\right\}}\right) & \\
& -K E\left(e^{-\int_0^T r_u(x) d u} \mathbf{1}_{\left\{r_T(x)<r^*\right\}}\right) .
\end{aligned}
$$

$$
E\left(e^{-\int_0^T r_u(x) d u} B\left(T, T^*\right)\right)=B\left(0, T^*\right), \quad E\left(e^{-\int_0^T r_u(x) d u}\right)=B(0, T) .
$$

Define two new probability measures $P_1$ and $P_2$ by, 

$$
\left.\frac{d P_1}{d P}\right|_{\mathcal{F}_T}=\frac{e^{-\int_0^T r_*(z) d u} B\left(T, T^*\right)}{B\left(0, T^*\right)},\left.\quad \frac{d P_2}{d P}\right|_{\mathcal{F}_T}=\frac{e^{-\int_0^T r_u(x) d u}}{B(0, T)} .
$$

$$
V_0=B\left(0, T^*\right) P_1\left(r_T(x)<r^*\right)-K B(0, T) P_2\left(r_T(x)<r^*\right) .
$$

$$
\begin{aligned}
& K_1=\frac{\delta^2}{2} \cdot \frac{e^{\gamma T}-1}{\gamma\left(e^{\gamma T}+1\right)+\left(\sigma^2 \psi_{0,1}\left(T^*-T\right)+b\right)\left(e^{\gamma T}-1\right)}, \\
& K_2=\frac{\sigma^2}{2} \cdot \frac{e^{\gamma T}-1}{\gamma\left(e^{\gamma T}+1\right)+b\left(e^{\gamma T}-1\right)} .
\end{aligned}
$$

Then it can be shown that the law of $\frac{r_r(x)}{K_1}$ under $P_1$ (resp. the law of $\frac{{ }^{T T}(x)}{K_2}$ under $P_2$ ) is a decentral chi-square with $\frac{4 a}{\sigma^2}$ degrees of freedom and decentral parameter $\xi_1$ (resp. $\xi_2$ ), 

$$
\xi_1=\frac{8 r_0(x) \gamma^2 e^{\gamma T}}{\sigma^2\left(e^{\gamma T}-1\right)\left(\gamma\left(e^{\gamma T}+1\right)\right)+\left(\sigma^2 \psi_{0,1}\left(T^*-T\right)+b\right)\left(e^{\gamma T}-1\right)},
$$

$$
\xi_2=\frac{8 r_0(x) \gamma^2 e^{\gamma T}}{\sigma^2\left(e^{\tau T}-1\right)\left(\gamma\left(e^{\gamma T}+1\right)+b\left(e^{\gamma T}-1\right)\right)} .
$$

\section*{Concepts of Navier-Stokes equations.}

The Navier-Stokes \footnote{\href{https://www.comsol.com/multiphysics/navier-stokes-equations}{COMSOL: Navier Stokes}} \footnote{\href{https://www.simscale.com/docs/simwiki/numerics-background/what-are-the-navier-stokes-equations/}{SIMSCALE: Navier Stokes}} equations are a set of partial differential equations that describe the motion of viscous fluids\cite{hamielec1967numerical},\cite{hamielec1967numerical}.

$$
\frac{\operatorname{Re}}{2}\left[\frac{\partial \Psi}{\partial r} \frac{\partial}{\partial \theta}\left(\frac{E^2 \Psi}{r^2 \sin ^2 \theta}\right)-\frac{\partial \Psi}{\partial \theta} \frac{\partial}{\partial r}\left(\frac{E^2 \Psi}{r^2 \sin ^2 \theta}\right)\right] \sin \theta=E^4 \Psi
$$

where $r$ is radius, $\theta$ is the polar angle, $\Psi$ is the stream function, $R e$ is the Reynolds number \cite{tang2020robust},\cite{thuerey2020deep}.  

$$
E^2=\frac{\partial^2}{\partial r^2}+\frac{\sin \theta}{r^2} \frac{\partial}{\partial \theta}\left(\frac{1}{\sin \theta} \frac{\partial}{\partial \theta}\right)
$$

\begin{figure}[H] 
\centering
\includegraphics[scale=0.27]{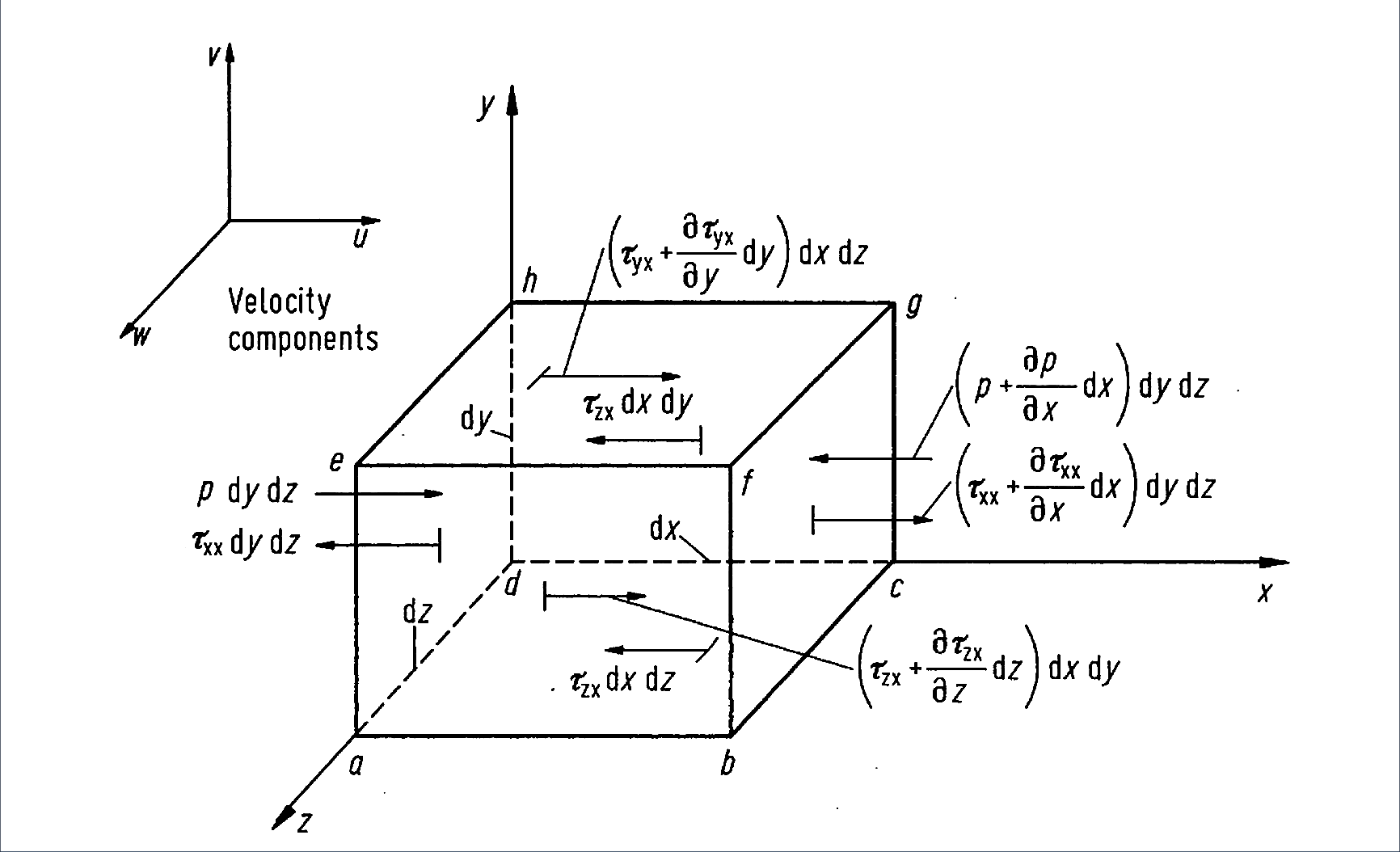} 
\caption{Navier-Stokes 3D.}
\label{fig:Navier_stokes_3D}
\end{figure}

The Reynolds number is the ratio of inertial forces to viscous forces:

$$
R e=\frac{\rho V D}{\mu}
$$

where $\rho$ is the density of the fluid, $V$ is the fluid velocity (Figure \ref{fig:Navier_stokes_3D}) at $r=\infty, D$ is the sphere diameter, and $\mu$ is the dynamic vicosity of the fluid. The boundary conditions for flow around a sphere are:

$$
\begin{array}{ll}
\Psi=\frac{1}{2} r^2 \sin ^2 \theta, & r \rightarrow \infty \\
\Psi=\frac{\partial \Psi}{\partial r}=0, & r=R
\end{array}
$$

The flow velocity field $v$ and pressure $\chi$ fulfill the incompressible Navier-Stokes equations \cite{mattingly1999elementary},\cite{foias2001navier},\cite{temam1995navier}.

$$
\begin{aligned}
\partial_t v+v \cdot \nabla v-\frac{1}{\operatorname{Re}} \Delta v+\nabla \chi & =f \\
\operatorname{div} v & =0
\end{aligned}
$$

on $Q_{\infty}:=\Omega \times(0, \infty)$ with a bounded and connected domain $\Omega \subseteq \mathbb{R}^d, d=2,3$, with boundary $\Gamma:=\partial \Omega$ of class $C^4$, a Dirichlet boundary condition $v=g$ on $\Sigma_{\infty}:=\Gamma \times(0, \infty)$, and appropriate initial conditions.

Now assume we are given a regular solution $w$ of the stationary Navier-Stokes \footnote{\href{https://physics.stackexchange.com/questions/513516/what-is-the-physical-meaning-of-navier-stokes-equations}{advection-diffusion : Navier Stokes equations}} equations, \cite{sohr2012navier},\cite{chorin1968numerical}, \cite{temam2024navier}.

$$
\begin{aligned}
w \cdot \nabla w-\frac{1}{\operatorname{Re}} \Delta w+\nabla \chi_s & =f \\
\operatorname{div} w & =0
\end{aligned}
$$

\section*{Concepts of Reinforcement learning.}

The value function for policy, 

$$
\begin{aligned}    
v_0^\pi(s)=\mathbb{E}\left[R_0\left(s_0, a_0\right)+\gamma R_1\left(s_1, a_1\right)+\ldots \mid s_0=s, \pi\right] \\ =\mathbb{E}\left[\sum_n \gamma^n R_n\left(s_n, a_n\right) \mid s_0=s, \pi\right]
\end{aligned}
$$

The Bellman equation for value function, 

$$
v_t^\pi(s)=R_t\left(s_t\right)+\gamma \sum_{s_{t+1} \in S} p\left(s_{t+1} \mid s_t, a_t=\pi\left(s_t\right)\right) v_{t+1}^\pi\left(s_{t+1}\right)
$$

The optimal value function: $v_t^*\left(s_t\right)=\max _\pi v_t^\pi\left(s_t\right)$, the Bellman optimality equation for optimal value function $v^*(s)$

$$
v_t^*\left(s_t\right)=R_t\left(s_t\right)+\max _{a_t \in \mathcal{A}} \gamma \sum_{s_{t+1} \in S} p\left(s_{t+1} \mid s_t, a_t\right) v_{t+1}^*\left(s_{t+1}\right)
$$

The optimal policy:

$$
\pi_t\left(s_t\right)=\underset{a_t \in \mathcal{A}}{\arg \max } \sum_{s_{t+1} \in S} p\left(s_{t+1} \mid s_t, a_t\right) v_t^*\left(s_{t+1}\right)
$$

$\begin{aligned}
\hspace{0.1cm} \nabla v_\pi(s) & =\nabla\left[\sum_a \pi(a \mid s) q_\pi(s, a)\right], \quad \text { for all } (s,a) \in \mathcal{S} \times \mathcal{A} \quad  \\
& =\sum_a\left[\nabla \pi(a \mid s) q_\pi(s, a)+\pi(a \mid s) \nabla q_\pi(s, a)\right] \quad  \\
& =\sum_a\left[\nabla \pi(a \mid s) q_\pi(s, a)+\pi(a \mid s) \nabla \sum_{s^{\prime}, \pi} p\left(s^{\prime}, r \mid s,a\right)\left(r+v_\pi\left(s^{\prime}\right)\right)\right]
\end{aligned}$

$=\sum_a\left[\nabla \pi(a \mid s) q_\pi(s, a)+\pi(a \mid s) \sum_{r^{\prime}} p\left(s^{\prime} \mid s, a\right) \nabla v_\pi\left(s^{\prime}\right)\right]$

$\begin{aligned}
\hspace{0.3cm} =  & \sum_a\left[\nabla \pi(a \mid s) q_\pi(s, a)+\pi(a \mid s) \sum_{s^{\prime}} p\left(s^{\prime} \mid s, a\right)\right. \\ & \left.\sum_{a^{\prime}}\left[\nabla \pi\left(a^{\prime} \mid s^{\prime}\right) q_\pi\left(s^{\prime}, a^{\prime}\right)+\pi\left(a^{\prime} \mid s^{\prime}\right) \sum_{s^{\prime \prime}} p\left(s^{\prime \prime} \mid s^{\prime}, a^{\prime}\right) \nabla v_\pi\left(s^{\prime \prime}\right)\right]\right]
\end{aligned}$

$$
=\sum_{x (s,a)\in \mathcal{S}\times \mathcal{A}} \sum_{k=0}^{\infty} \operatorname{Pr}(s \rightarrow x, k, \pi) \sum_a \nabla \pi(a \mid x) q_\pi(x, a)
$$

Where $\operatorname{Pr}(s \rightarrow x, k, \pi)$ is the probability of transitioning from state $s$ to state $x$ in $k$ steps under policy $\pi$.

$$
\begin{aligned}
\nabla J(\boldsymbol{\theta}) & =\nabla v_\pi\left(s_0\right) \\
& =\sum_s\left(\sum_{k=0}^{\infty} \operatorname{Pr}\left(s_0 \rightarrow s, k, \pi\right)\right) \sum_a \nabla \pi(a \mid s) q_\pi(s, a) \\
& =\sum_s \eta(s) \sum_a \nabla \pi(a \mid s) q_\pi(s, a) \\
& =\sum_{s^{\prime}} \eta\left(s^{\prime}\right) \sum_s \frac{\eta(s)}{\sum_{s^{\prime}} \eta\left(s^{\prime}\right)} \sum_a \nabla \pi(a \mid s) q_\pi(s, a) \\
& =\sum_{s^{\prime}} \eta\left(s^{\prime}\right) \sum_s \mu(s) \sum_a \nabla \pi(a \mid s) q_\pi(s, a) \\
& \propto \sum_s \mu(s) \sum_a \nabla \pi(a \mid s) q_\pi(s, a)
\end{aligned}
$$

$$
\begin{aligned}
\hspace{0.1cm} \nabla v_\pi(s) & = \nabla\left[\sum_a \pi(a \mid s) q_\pi(s, a)\right], \quad \text { for all } s \in S \\
 & = \sum_a\left[\nabla \pi(a \mid s) q_\pi(s, a)+\pi(a \mid s) \nabla q_\pi(s, a)\right] \quad  \\
& \hspace{-0.9cm} =  \sum_a\left[\nabla \pi(a \mid s) q_\pi(s, a)+\pi(a \mid s) \nabla \sum_{s^{\prime}, r} p\left(s^{\prime}, r \mid s, a\right)\left(r-r(\theta)+v_\pi\left(s^{\prime}\right)\right)\right] \\
&  \hspace{-0.5cm} =  \sum_a\left[\nabla \pi(a \mid s) q_\pi(s, a)+\pi(a \mid s)\left[-\nabla r(\theta)+\sum_{x^{\prime}} p\left(s^{\prime} \mid s, a\right) \nabla v_\pi\left(s^{\prime}\right)\right]\right]
\end{aligned}
$$

$$
\hspace{0.1cm} \nabla r(\boldsymbol{\theta}) =\sum_a\left[\nabla \pi(a \mid s) q_\pi(s, a)+\pi(a \mid s) \sum_{s^{\prime}} p\left(s^{\prime} \mid s, a\right) \nabla v_\pi\left(s^{\prime}\right)\right]-\nabla v_\pi(s)
$$

$$
\begin{aligned}
 \hspace{-9mm}\nabla J(\theta)= & \sum_s \mu(s)\left(\sum_a\left[\nabla \pi(a \mid s) q_\pi(s, a)+\pi(a \mid s) \sum_{s^{\prime}}  p\left(s^{\prime} \mid s, a\right) \nabla v_\pi\left(s^{\prime}\right)\right]-\nabla v_\pi(s)\right) \\
\hspace{-4mm} =  & \sum_s \mu(s) \sum_a \nabla \pi(a \mid s) q_\pi(s, a) \\& \quad+\sum_s \mu(s) \sum_a \pi(a \mid s) \sum_{s^{\prime}} p\left(s^{\prime} \mid s, a\right) \nabla v_\pi\left(s^{\prime}\right)-\sum_s \mu(s) \nabla v_\pi(s)
\end{aligned}
$$

$$
\begin{aligned}
= & \sum_s \mu(s) \sum_a \nabla \pi(a \mid s) q_\pi(s, a) \\
& \quad+\sum_{s^{\prime}} \underbrace{\sum_s \mu(s) \sum_a \pi(a \mid s) p\left(s^{\prime} \mid s, a\right)}_{\mu\left(s^{\prime}\right)} \nabla v_\pi\left(s^{\prime}\right)-\sum_s \mu(s) \nabla v_\pi(s) \\
= & \sum_s \mu(s) \sum_a \nabla \pi(a \mid s) q_\pi(s, a)+\sum_{s^{\prime}} \mu\left(s^{\prime}\right) \nabla v_\pi\left(s^{\prime}\right)-\sum_n \mu(s) \nabla v_\pi(s) \\
= & \sum_s \mu(s) \sum_a \nabla \pi(a \mid s) q_\pi(s, a) . \quad 
\end{aligned}
$$

\section*{Concepts of stochastic volatility jump.}

$$
d S_t=\left(\mu_S+\lambda j\right) S_t d t+\sqrt{v_t} S_t d B_t-S_t j d N_t
$$
$0 \leq j<1$, $\lambda=0$ or the jump size $j=0$, the number of jumps between $t$ and $t+d t$. Applying Ito's lemma\footnote{\href{https://www.lexifi.com/blog/quant/jump-diffusion-models-merton-and-bates/}{ LexiFi: Bates Model}} for semi-martingales \cite{eisenberg2022ito}, \cite{guo2023ito}, \cite{yan2006option}.

$$
d \ln S_t=\left(\mu_S-\frac{1}{2} v_t+\lambda j\right) d t+\sqrt{v_t} d B_t+\ln (1-i) d N_t
$$

$$
\ln S_{k+1}=\ln S_k+\left(\mu_S-\frac{1}{2} v_k+\lambda j\right) \Delta t+\sqrt{v_t} \sqrt{\Delta t} B_k+\mu_k
$$

 $\mu_0=0$ , 

$$
\mu_k=\delta_0(0) e^{-\lambda \Delta t}+\delta_0(\ln (1-j))\left(1-e^{-\lambda \Delta t}\right)
$$

$\delta_0(0)$ corresponds to the Dirac $\delta$  function.

\begin{theorem}

The rational price of a standard European call option \footnote{\href{https://math.stackexchange.com/questions/900911/proof-of-the-black-scholes-pricing-formula-for-european-call-option}{Full Proof: European call option}} is, 

$$
C\left(T,\left(K-S_T^1\right)^{+}\right)=S_0^1 \Phi\left(d_{+}\right)-K e^{-r T} \Phi\left(d_{-}\right) .
$$

Where $\Phi(y)=\frac{1}{\sqrt{2 \pi}} \int_{-\infty}^y e^{-\frac{1}{2} z^2} d z$ is the standard normal cumulative distribution function, 

$$
d_{+}=\frac{\log \left(\frac{S_0^1}{K}\right)+T\left(r+\frac{\sigma^2}{2}\right)}{\sigma \sqrt{T}}, \quad d_{-}=\frac{\log \left(\frac{S_0^1}{K}\right)+T\left(r-\frac{\sigma^2}{2}\right)}{\sigma \sqrt{T}} .
$$

($d_{=} d_{+}-\sigma \sqrt{T}$.)
The minimal hedge $\phi_t^*=\left(H_t^0, H_t^1\right)$,

$$
\begin{aligned}
& H_t^1=\Phi\left(\frac{\log \left(\frac{S_t^1}{K}\right)+(T-t)\left(r+\frac{\sigma^2}{2}\right)}{\sigma \sqrt{T-t}}\right) \\
& H_t^0=-e^{-r T} K \Phi\left(\frac{\log \left(\frac{S_t^1}{K}\right)+(T-t)\left(r-\frac{\sigma^2}{2}\right)}{\sigma \sqrt{T-t}}\right)
\end{aligned}
$$

$$
V_t\left(\phi^*\right)=H_t^0 S_t^0+H_t^1 S_t^1
$$

\end{theorem}

\vspace{2cm}

\begin{proof}

$f(s)=(s-K)^{+}$, 

$$
\begin{aligned}
F(t, s) & =\frac{1}{\sqrt{2 \pi}} \int_{-\infty}^{\infty} f\left(s \exp \left\{\sigma y \sqrt{t}+\left(r-\frac{\sigma^2}{2}\right) t\right\}\right) e^{-\frac{1}{2} y^2} d y \\
& =\frac{1}{\sqrt{2 \pi}} \int_{y(t, s)}^{\infty}\left(s \exp \left\{\sigma y \sqrt{t}+\left(r-\frac{\sigma^2}{2}\right) t\right\}-K\right) e^{-\frac{1}{2} y^2} d y
\end{aligned}
$$

where $y(t, s)$ is the solution of,

$$
s \exp \left\{\sigma y \sqrt{t}+\left(r-\frac{\sigma^2}{2}\right) t\right\}=K
$$

$$
y(t, s)=\sigma^{-1} t^{-\frac{1}{2}}\left(\log \left(\frac{K}{s}\right)-\left(r-\frac{\sigma^2}{2}\right) t\right)
$$

$$
\begin{aligned}
F(t, s) & =\frac{e^{r t}}{\sqrt{2 \pi}} \int_{y(t, s)}^{\infty} s \exp \left\{\sigma y \sqrt{y}-\sigma^2 \frac{t}{2}-\frac{1}{2} y^2\right\} d y-K[1-\Phi(y(t, s))] \\
& =\frac{s e^{r t}}{\sqrt{2 \pi}} \int_{y(t, s)-\sigma \sqrt{t}}^{\infty} e^{-\frac{1}{2} x^2} d x-K[1-\Phi(y(t, s))] \\
& =s e^{r t}[1-\Phi(y(t, s)-\sigma \sqrt{t})]-K[1-\Phi(y(t, s))]
\end{aligned}
$$

$$
\begin{aligned}
C\left(T,\left(S_T-K\right)^{+}\right) & =e^{-r T} F\left(T, S_0\right) \\
& =S_0 \Phi\left(\sigma \sqrt{T}-y\left(T, S_0\right)\right)-K e^{-r T} \Phi\left(-y\left(T, S_0\right)\right) \\
& =S_0 \Phi\left(d_{+}\right)-K e^{-r T} \Phi\left(d_{-}\right) .
\end{aligned}
$$

The minimal hedge is $H_t^1=e^{-r(T-t)} \frac{\partial F}{\partial s}\left(T-t, S_t\right)$ , 

$$
\begin{aligned}
H_t^1 & =\Phi\left(\sigma \sqrt{T-t}-y\left(T-t, S_t\right)\right) \\
& =\Phi\left(\sigma \sqrt{T-t}-\sigma^{-1}(T-t)^{-\frac{1}{2}}\left(\log \left(\frac{K}{S_t}\right)-\left(r-\frac{\sigma^2}{2}\right)(T-t)\right)\right) \\
& =\Phi\left(\frac{\log \left(\frac{S_s}{K}\right)+(T-t)\left(r+\frac{\sigma^2}{2}\right)}{\sigma \sqrt{T-t}}\right)
\end{aligned}
$$

$$
\begin{aligned}
V_t\left(\phi^*\right)= & e^{-r(T-t)} F\left(T-t, S_t\right) \\
= & S_t \Phi\left(\frac{\log \left(\frac{S_t}{K}\right)+(T-t)\left(r+\frac{\sigma^2}{2}\right)}{\sigma \sqrt{T-t}}\right) \\
& -K e^{-r(T-t)} \Phi\left(\frac{\log \left(\frac{S_t}{K}\right)+(T-t)\left(r-\frac{\sigma^2}{2}\right)}{\sigma \sqrt{T-t}}\right) .
\end{aligned}
$$

$$
\begin{aligned}
H_t^0 & =e^{-r t} V_t\left(\phi^*\right)-e^{-r t} H_t^1 S_t \\
& =-K e^{-r T} \Phi\left(\frac{\log \left(\frac{S_t}{K}\right)+\left(r-\frac{\sigma^2}{2}\right)(T-t)}{\sigma \sqrt{T-t}}\right)
\end{aligned}
$$.

\end{proof}

\section*{Ablation study: study of an alternative detection method.}

Using a Graph RNN autoencoder methodology and clustering techniques, this section suggests a novel method for precisely identifying and mitigating backdoor risks in audio data. We suggest a Graph RNN autoencoder that effectively detects backdoor attacks by learning a low-dimensional latent-space representation of the input data while maintaining the graph structure. Through fine-tuning and retraining on a mixed dataset, our method enhances the model's resilience by combining reconstruction loss-based detection utilizing an autoencoder with K-means clustering, PCA (Principal Component Analysis), and LOF (Local Outlier Factor).

\subsection{Proposed Method.}  

Deep learning models, such Graph and Recurrent Neural Networks (Graph RNNs) \cite{xu2023rethinking},\cite{xu2022more}, have been applied in a number of domains, such as anomaly detection, natural language processing, and picture categorization. Backdoor attacks, however, have the potential to compromise these models' security and performance. In this study, we provide a strong Graph RNN \footnote{\href{https://gnn.seas.upenn.edu/lecture-11/}{GRNN}} autoencoder method to boost resilience and strengthen defenses against backdoor attacks.

\subsection{Backdoor Attack Detection: clustering algorithms.}

We use two complementing methods to detect backdoor attack cases: K-means clustering using principal component analysis (PCA) and local outlier factor (LOF) and reconstruction loss-based detection using an autoencoder.

\vspace{0.3cm}

The autoencoder was trained on a dataset that included both poisoned and benign samples for reconstruction loss-based detection. By measuring the difference between the input data and its reconstruction, the reconstruction loss, represented by $\mathcal{L}_\text{recon}$, was computed: $\mathcal{L}_\text{recon} = \|\mathbf{X} - \mathbf{X}'\|^2$. The reconstruction loss of benign samples was used to estimate an anomaly detection threshold, represented by $\tau$.

\vspace{0.3cm}

The input samples were flattened to create a feature matrix $\mathbf{F}$ for K-means clustering with PCA and LOF. The feature matrix's dimensionality was then decreased using Principal Component Analysis (PCA): $\mathbf{F}_{\text{reduced}} = \text{PCA}(\mathbf{F})$. To find clusters, the reduced feature matrix was subjected to K-means clustering. Using the K-means technique, the cluster centroids, represented by $\mathbf{C}$, were determined. The formula for calculating the Euclidean distance between samples and the cluster centroids is $\mathbf{D} = \|\mathbf{F}_{\text{reduced}} - \mathbf{C}\|$. After that, anomalies in the distance measurements are found using the local outlier factor (LOF).

\subsection{Fine-tuning and Retraining.}

To evaluate the effectiveness of the developed strategy, experiments were performed on a dataset composed of benign samples and samples contaminated by backdoor attacks. For this purpose, the dataset was divided into training and testing sets. The autoencoder was first trained on the training set and then used for detection on the test set. Finally, the detection accuracy was calculated by comparing the predicted labels with the ground-truth labels.

\subsection{Detection Accuracy.}

The detection accuracy of the autoencoder was measured by comparing the reconstruction loss of the test samples with the anomaly detection threshold. Let $\mathbf{L}_\text{test}$ represent the reconstruction loss of the test samples. The backdoor instances are identified as:

$\text{BD}_\text{autoencoder} = (\mathbf{L}_\text{test} > \tau)$, where $\tau$ is the threshold determined for the training set.

The detection accuracies of K-means clustering with PCA and the LOF method were also calculated. Let $\mathbf{A}_\text{Kmeans-LOF}$ represent the anomaly scores obtained by using this method. The backdoor instances are identified as: 

$\text{BD}_\text{Kmeans-LOF} = (\mathbf{A}_\text{Kmeans-LOF} > \tau)$.
The detection accuracy was then computed as the ratio of correctly identified backdoor instances to the total number of backdoor instances in the test set.


\subsubsection{3D visualisation of GRNN-AE.}
  
\begin{figure}
  \begin{minipage}[b]{0.24\textwidth}
    \includegraphics[width=\textwidth]{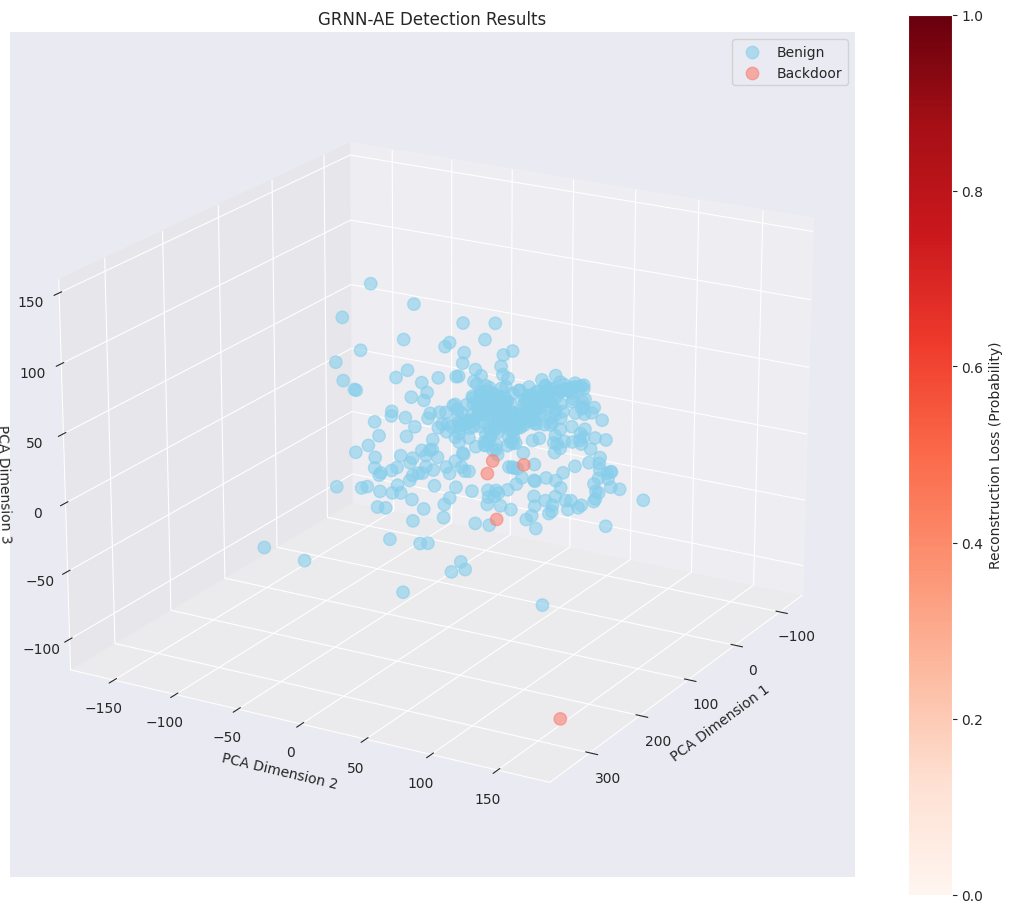} 
    \caption{GRNN-AE.}
    \label{fig:ii}
  \end{minipage}
  \hfill
  \begin{minipage}[b]{0.24\textwidth}
    \includegraphics[width=\textwidth]{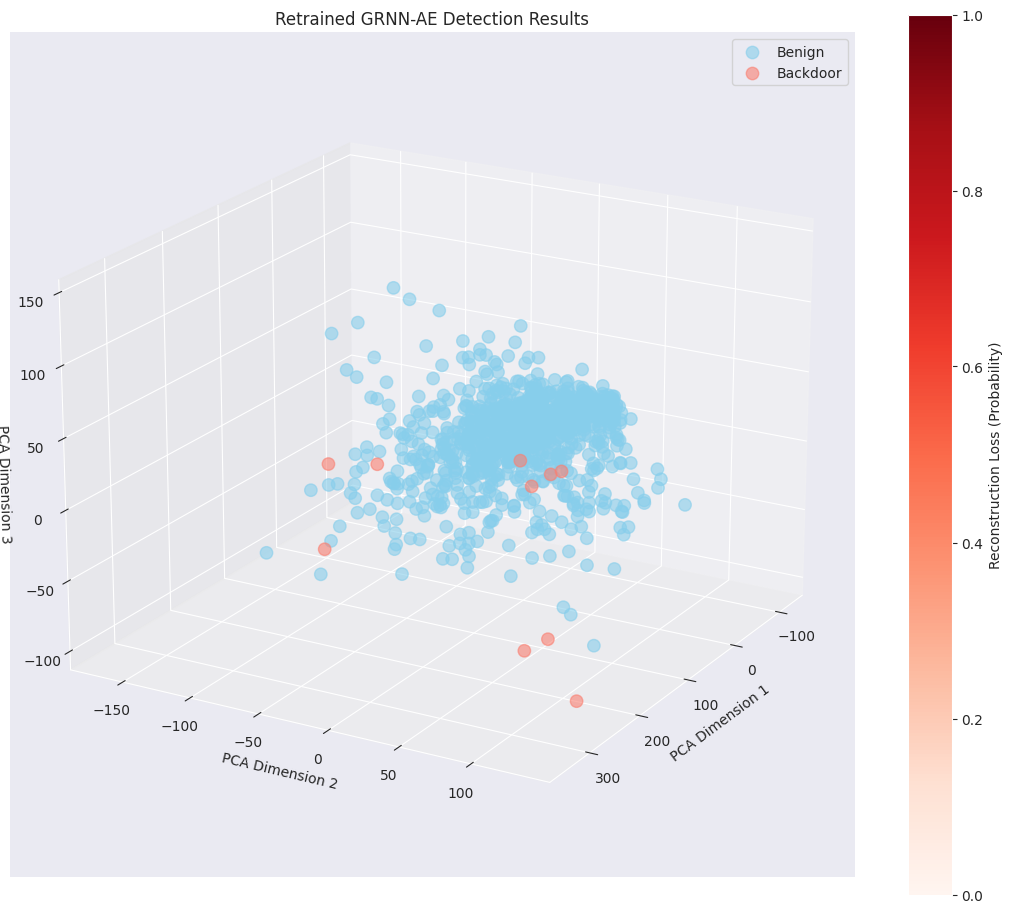}
    \caption{Retrained GRNN-AE.}
    \label{fig:j}
  \end{minipage}
\end{figure}

We illustrate a 3D scatterplot visualization. Figure \ref{fig:ii}, Figure \ref{fig:j} of the outcomes of the autoencoding method for backdoor attack detection, in a condensed three-dimensional space computed using PCA, the graphic emphasizes the distinction between benign and possible poisoned samples. When compared to a model that has been retrained, the comparison of color-coded probability scores and detection accuracy data shows the impact of retraining on the performance of backdoor detection (GRNN-AE).

\bibliographystyle{IEEEtran}

\bibliography{IEEEabrv,refs}

\end{document}